\documentclass[lettersize,journal]{IEEEtran}
\usepackage[caption=false,font=normalsize,labelfont=sf,textfont=sf]{subfig}
\usepackage{stfloats}
\usepackage{verbatim}
\usepackage{cite}
\usepackage{amsmath,amssymb,amsfonts}
\usepackage{graphicx}
\usepackage{algorithm}

\usepackage{algpseudocode}
\usepackage{threeparttable}
\usepackage{graphics}
\usepackage{epsfig}
\usepackage{textcomp}
\usepackage{xcolor}
\usepackage{tcolorbox}
\usepackage{amssymb}
\usepackage{booktabs}
\usepackage{multirow}
\usepackage{makecell}
\usepackage{bbding}  
\usepackage{pifont}
\usepackage{multicol}
\usepackage{hyperref}
\usepackage{colortbl}
\definecolor{light-blue}{HTML}{0095d9}
\definecolor{dark-green}{HTML}{006400}
\definecolor{dark-orange}{HTML}{D1480E}
\definecolor{dark-blue}{HTML}{1976D2}
\definecolor{dark-purple}{HTML}{8d4bbb}
\definecolor{dark-red}{HTML}{D63C3C}
\definecolor{n1}{HTML}{ff9999}
\definecolor{n2}{HTML}{FFCC99}
\definecolor{n3}{HTML}{FFFF99}

\hypersetup{
    colorlinks=true,
    linkcolor=dark-red,
    filecolor=magenta,      
    urlcolor=dark-red,
    citecolor=dark-blue,
    pdfpagemode=FullScreen,
}
\usepackage{url}
\usepackage{mathtools}

\usepackage{enumitem}
\setlist[itemize]{leftmargin=4mm}
\usepackage{amsmath}
\usepackage{amsfonts}
\usepackage{wrapfig}
\usepackage{booktabs} % for professional tables
\usepackage{pifont}

\usepackage[normalem]{ulem}
\usepackage{multirow}
\usepackage[algo2e]{algorithm2e}
\SetKwComment{Comment}{$\triangleright$\ }{}
\SetKwComment{CommentL}{$\/\/$}{}
\SetKwFunction{FMain}{Main}
\usepackage{tikz}
\usetikzlibrary{fit,calc}

\usepackage{amssymb}% http://ctan.org/pkg/amssymb
\usepackage{pifont}% http://ctan.org/pkg/pifont
\newcommand{\cmark}{\ding{51}}%
\newcommand{\xmark}{\ding{55}}%
\newcommand*{\tikzmka}[1]{\tikz[remember picture,overlay,] \node (#1) {};\ignorespaces}
\newcommand{\boxita}[1]{\tikz[remember picture,overlay]{\node[yshift=3pt,fill=#1,opacity=.15,fit={($(A)+(0.005\linewidth,0.2\baselineskip)$)($(B)+(0.02\linewidth,-.2\baselineskip)$)}] {};}\ignorespaces}
\newcommand{\boxitc}[1]{\tikz[remember picture,overlay]{\node[yshift=3pt,fill=#1,opacity=.1,fit={($(A)+(0.005\linewidth,0.2\baselineskip)$)($(B)+(0.02\linewidth,-0.2\baselineskip)$)}] {};}\ignorespaces}
\colorlet{pink}{red!40}
\colorlet{cyan}{cyan!60}
\colorlet{orange}{orange!80}

\newcommand{\xy}[1]{\textcolor{black}{#1}}
\newcommand{\shh}[1]{\textcolor{black}{#1}}
\newcommand{\huihong}[1]{\textcolor{black}{#1}}
\newcommand*{\tikzmkc}[1]{\tikz[remember picture,overlay,] \node (#1) {};\ignorespaces}

\begin{document}

\title{NASH: Neural Architecture and Accelerator Search for Multiplication-Reduced Hybrid Models}

\author{Yang Xu, Huihong Shi, and Zhongfeng Wang,~\IEEEmembership{Fellow,~IEEE}

\thanks{This work was supported by the National Key R\&D Program of China under Grant 2022YFB4400604.}
\thanks{Yang Xu and Huihong Shi contributed equally to this work, and are with the School of Electronic Science and Engineering, Nanjing University, Nanjing, China (e-mail: \{{xyang, shihh}\}@smail.nju.edu.cn).}
\thanks{Zhongfeng Wang is with the School of Electronic Science and Engineering, Nanjing University, and the School of Integrated Circuits, Sun Yat-sen University (email: zfwang@nju.edu.cn).}
\thanks{Correspondence should be addressed to Zhongfeng Wang.}}

\maketitle
\begin{abstract}
\textcolor{black}{
% While multiplication-free models are well-known for their significantly reduced latency and energy consumption, they typically exhibit lower accuracy compared to their multiplication-based counterparts. Therefore, prior works have built multiplication-reduced hybrid models via Neural Architecture Search (NAS) to maintain accuracy while reducing hardware costs, i.e., NASA and NASA-F. However, on one hand, they all overlook the exploration of accelerator search and rely on handcrafted accelerators, resulting in lower hardware performance. On the other hand, they either require annoying retraining or suffer from gradient conflict issues, thus getting unsatisfactory algorithm results. In this paper, we propose Hybrid-NAAS, a Neural Architecture and Accelerator Search (NAAS) framework dedicated to multiplication-reduced hybrid models. Specifically, on the \underline{architecture search level}, we leverage zero-shot metrics to search for potentially good sub-networks in advance, and then allocate training resources to optimize the selected ones instead of randomly sampled ones to boost accuracy. On the \underline{accelerator search level}, we simultaneously search accelerator architecture and mapping strategy, exploiting the hardware potential of our hybrid models to enhance their throughput. Overall, Hybrid-NAAS explores both the architecture and accelerator search space to become the first NAAS framework for multiplication-reduced hybrid models. Extensive experimental results...
\shh{
% Multiplication is the most costly operation in the computation of modern deep neural networks (DNNs), challenging DNNs' deployment on edge devices. To enhance hardware efficiency, multiplication-free models have been proposed, which, however, exhibit lower accuracy compared to their multiplication-based counterparts. Consequently, multiplication-reduced hybrid models have emerged to marry the benefits of both multiplication-based and multiplication-free models. 
The significant computational cost of multiplications hinders the deployment of deep neural networks (DNNs) on edge devices. While multiplication-free models offer enhanced hardware efficiency, they typically sacrifice accuracy. As a solution, multiplication-reduced hybrid models have emerged to combine the benefits of both approaches.
Particularly, prior works, i.e., NASA and NASA-F, leverage Neural Architecture Search (NAS) to construct such hybrid models, enhancing hardware efficiency while maintaining accuracy.
However, they either entail costly retraining or encounter gradient conflicts, limiting both search efficiency and accuracy. 
Additionally, they overlook the acceleration opportunity introduced by accelerator search, yielding sub-optimal hardware performance. 
To overcome these limitations, we propose NASH, a \textit{N}eural architecture and \textit{A}ccelerator \textit{S}earch framework for multiplication-reduced \textit{H}ybrid models. Specifically, as for \underline{NAS}, we propose a {tailored zero-shot metric} to pre-identify promising \xy{hybrid} models before training, enhancing search efficiency while alleviating gradient conflicts. Regarding \underline{accelerator search}, 
we innovatively introduce {coarse-to-fine search} to streamline search process. 
\xy{Furthermore, we seamlessly integrate these two levels of searches to unveil NASH,}
obtaining optimal model and accelerator pairing.
%Experiments validate our effectiveness, e.g., $\uparrow$$0.56\%$ accuracy on Tiny-ImageNet over prior multiplication-reduced work NASA-F, and $\uparrow$$2.14\times$ throughput and $\uparrow$$2.01\times$ FPS with $\uparrow$$0.25\%$ accuracy against the state-of-the-art multiplication-based system on CIFAR100.
{Experiments validate our effectiveness, e.g., when compared with the state-of-the-art multiplication-based system, we can achieve $\uparrow$$2.14\times$ throughput and $\uparrow$$2.01\times$ FPS with $\uparrow$$0.25\%$ accuracy on CIFAR-100, and $\uparrow$$1.40\times$ throughput and $\uparrow$$1.19\times$ FPS with $\uparrow$$0.56\%$ accuracy on Tiny-ImageNet.}
%Codes will be released upon acceptance.
Codes are available at \url{https://github.com/xuyang527/NASH.}
}}

\end{abstract}


\section{Introduction}
\label{sec:intro}
Despite the remarkable success of deep neural networks (DNNs) in various computer vision tasks \cite{He2016DeepRL}, \cite{GoingDW}, \cite{YOLO9000BF}, the involved intensive multiplications yield significant hardware costs, thus hindering DNNs' deployment on resource-constrained edge devices. To enhance hardware efficiency, prior works \cite{AdderNet}, \cite{AdderNetv2}, \cite{AdderTransformer}, \cite{DeepShift}, \cite{ShiftAddNet} have developed multiplication-free models that substitute multiplications with hardware-friendly operations, such as bitwise shifts and additions. 
%For example, DeepShift \cite{DeepShift} and AdderNet \cite{AdderNet} substitute multiplications with bitwise shifts and additions, respectively, thereby creating models that consist exclusively of shift layers or adder layers. 
For example, DeepShift \cite{DeepShift} substitutes multiplications with bitwise shifts, thereby building models exclusively with shift layers. Besides, AdderNet \cite{AdderNet} trades multiplications with additions, thus constructing models with solely adder layers.
Furthermore, ShiftAddNet \cite{ShiftAddNet} integrates both shift and adder layers, developing models with merely hardware-friendly bitwise shifts and additions. Despite their notable hardware efficiency, multiplication-free models are generally inferior to their multiplication-based counterparts in accuracy \cite{NASA+,nasa-f}. Thus, motivated by the high accuracy of multiplication-based models and the promising hardware efficiency of multiplication-free models, there is an urgent demand for multiplication-reduced hybrid models to marry the benefits of both multiplications and hardware-friendly operations.

\begin{figure}[t]
	\centerline{\includegraphics[width=\linewidth]{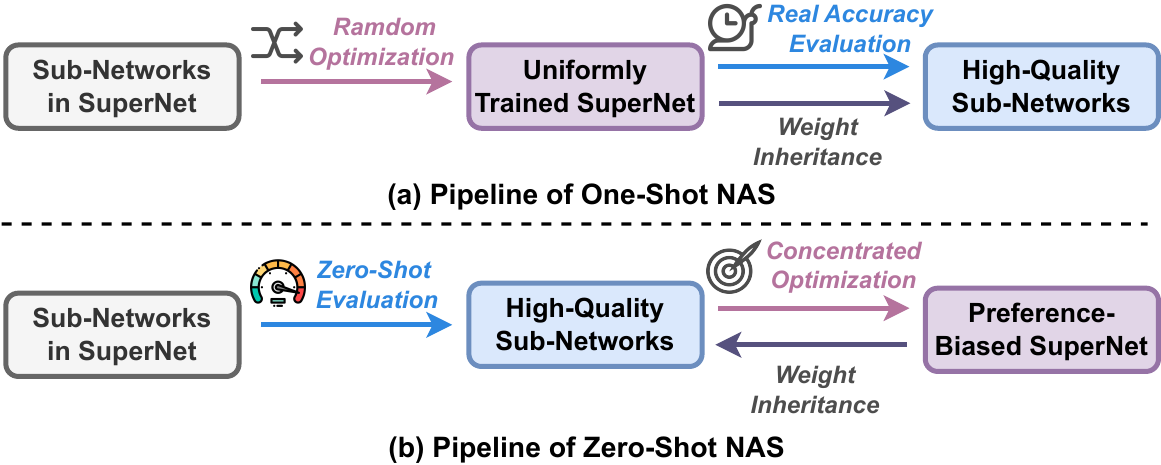}}
	\caption{Pipelines of (a) one-shot supernet-based NAS \cite{BigNAS}, \cite{AlphaNet} and (b) zero-shot NAS framework \xy{combined with preference-biased supernet training} \cite{wang2023prenas}.}
	\label{fig:nas_pipeline}
\end{figure}

To achieve this goal, prior works leverage Neural Architecture Search (NAS) 
%, which searches for the optimal network architecture in an enormous search space, 
to automatically design such multiplication-reduced hybrid models \cite{nasa-f,NASA,NASA+}. Specifically, NASA \cite{NASA} proposes a dedicated differentiable NAS (DNAS) engine to 
%relax the discrete search space to be continuous and 
automatically construct hybrid models in a differentiable manner, which yet requires an expensive retraining process for searched models. 
%However, the DNAS algorithm requires an expensive retraining/finetuning process to achieve ideal accuracy and needs to restart its search process once the resource constraints (e.g., model size) change.
%However, despite the success of DNAS, it suffers from several drawbacks. Specifically, (i) the networks discovered by the DNAS algorithm require an expensive retraining/finetuning process to achieve their optimal accuracy. (ii) a new iteration of the DNAS process is necessary whenever the resource constraints (e.g., model size, latency, or energy) change, causing additional search overhead. 
%To mitigate these limitations, 
Subsequently, NASA-F \cite{nasa-f} develops a tailored one-shot NAS method
% \textcolor{brown}{(whose pipeline generally follows the example in Fig. \ref{fig:nas_pipeline} (a))} 
to fully optimize all sub-networks within the pre-defined multiplication-reduced hybrid supernet, thus achieving high accuracy without retraining/finetuning. 
% Unlike in DNAS, sub-networks in one-shot NAS can directly inherit their parameters from the well-optimized supernet without retraining/finetuning. %and sub-networks that satisfy different resource constraints can be found through a resource-constrained architecture search process after supernet training. 
Despite its effectiveness, one-shot NAS still suffers from several drawbacks. Specifically, 
%TODO:
\shh{as shown in Fig. \ref{fig:nas_pipeline} (a), (i) one-shot NAS typically involves the random sampling of sub-networks with diverse qualities during the supernet training stage, thus yielding gradient conflicts and hindering the achievable accuracy; 
Additionally, (ii) the accuracy evaluation through model inference is time-consuming, thereby decelerating the subsequent architecture search process.}
% \xy{(i) random sampling from all sub-networks leads to a uniform distribution of training resources; (ii) Concurrently optimizing such a large population of sub-networks, each with varying quality, introduces gradient conflicts that impede the attainable accuracy; (iii) The real accuracy evaluation is time-consuming and thus decelerates the whole process.}
%\huihong{(i) gradient conflict issues could happen since one-shot NAS concurrently optimizes the high-quality and low-quality sub-networks. (ii) Evaluating involved in the resource-constrained architecture search process is time-consuming and thus decelerates the search.} 
To solve these limitations, zero-shot metrics have been \xy{applied} to predict promising networks before model optimization \cite{lee2018snip}, \cite{zen-nas}, \cite{TE-NAS}, \cite{grasp}, \cite{synflow}, \cite{fisher}.
For example, the recent work PreNAS \cite{wang2023prenas} 
% \textcolor{brown}{(as illustrated in Fig. \ref{fig:nas_pipeline} (b))} 
advocates using SNIP \cite{lee2018snip} to pre-identify potentially high-quality sub-networks before supernet training. 
% \textcolor{brown}{These identified sub-networks are then optimized during supernet training. Finally, they inherit the supernet's weights to serve as high-quality ready-to-use networks}.
%TODO:
% \textcolor{brown}{By doing this, (i) training resources can be concentrated on the selected promising sub-networks; (ii) gradient conflict can be alleviated since only high-quality networks are optimized; (iii) evaluation of real accuracy is replaced by zero-shot metrics, thus the search efficiency is enhanced.}
\shh{By doing this, as illustrated in Fig. \ref{fig:nas_pipeline} (b), (i) assessment using zero-shot metrics has been proven to be much faster than accuracy measurement \cite{wang2023prenas}, thus search efficiency is significantly enhanced; Besides, (ii) during the subsequent supernet training phase, 
%描述这个supernet training的两个地方（这里还有related works）里面都只是说 allocates training resources to ... 这种比较high level的， 感觉提一下这个具体的过程，也就是optimize selected net-works比较好
{training resources can be concentrated on these selected sub-networks instead of randomly sampled ones, allowing for the alleviation of gradient conflicts.}}
%By doing this, (i) training resources can be concentrated on the selected promising sub-networks to alleviate gradient conflict, thus boosting accuracy. (ii) Evaluation of real accuracy is replaced by zero-shot metrics, largely reducing the search time consumption. 
However, due to the different algorithmic properties between multiplication-based convolutions and multiplication-free operations \cite{AdderNet, NASA}, existing zero-shot metrics developed for multiplication-based models \cite{lee2018snip,gradnorm,synflow,grasp} are not directly applicable to our desired multiplication-reduced models, calling for the exploration of tailored ones.   

\shh{In parallel, various works \cite{Cai2019OnceFA, NASA+, nasa-f} have developed dedicated accelerators to improve the hardware efficiency of DNNs from the hardware perspective. However, considering the intricate design space of accelerators, including the accelerator configuration (e.g., PE number and buffer size) and mapping method (also dubbed dataflow), it is non-trivial to handcraft an optimal dedicated accelerator, calling for automatic tools \cite{NAAS}, \cite{DNA}, \cite{hong2023dosa}, \cite{anas}, \cite{EfficientCosearch}.
Specifically, NAAS \cite{NAAS} utilizes an evolutionary algorithm to search for the optimal accelerator configuration and mapping method. Then, it seamlessly integrates the accelerator search process with neural architecture search, thereby directly yielding the optimal pairing of model and accelerator.
Inspired by its success, it is highly desired to leverage accelerator search to automatically build the accelerator dedicated to our desired multiplication-reduced hybrid model for boosting hardware efficiency.
% However, due to the existence of heterogeneous layers within the hybrid model, including convolutions, shift layers, and adder layers, the design space of its tailored accelerator is substantially expanded \cite{NASA, nasa-f}. This is typically attributed to the need for distinct computing engines to independently accelerate various types of layers within hybrid models  \cite{NASA, nasa-f}, which inevitably impedes the pace of our accelerator search process and thus calls for fast search strategies.
However, due to the distinct algorithmic properties exhibited by heterogeneous layers within the hybrid model, including convolution {layers}, shift layers, and adder layers, there arises a necessity to construct separate computing engines for each type of layer, aiming to enable independent processing \cite{NASA, nasa-f}. This inherently expands the search {space} and inevitably amplifies the search complexity of our dedicated accelerator \cite{NASA}, thus necessitating fast search strategies.
}

\begin{figure*}[t]
	\centerline{\includegraphics[width=\linewidth]{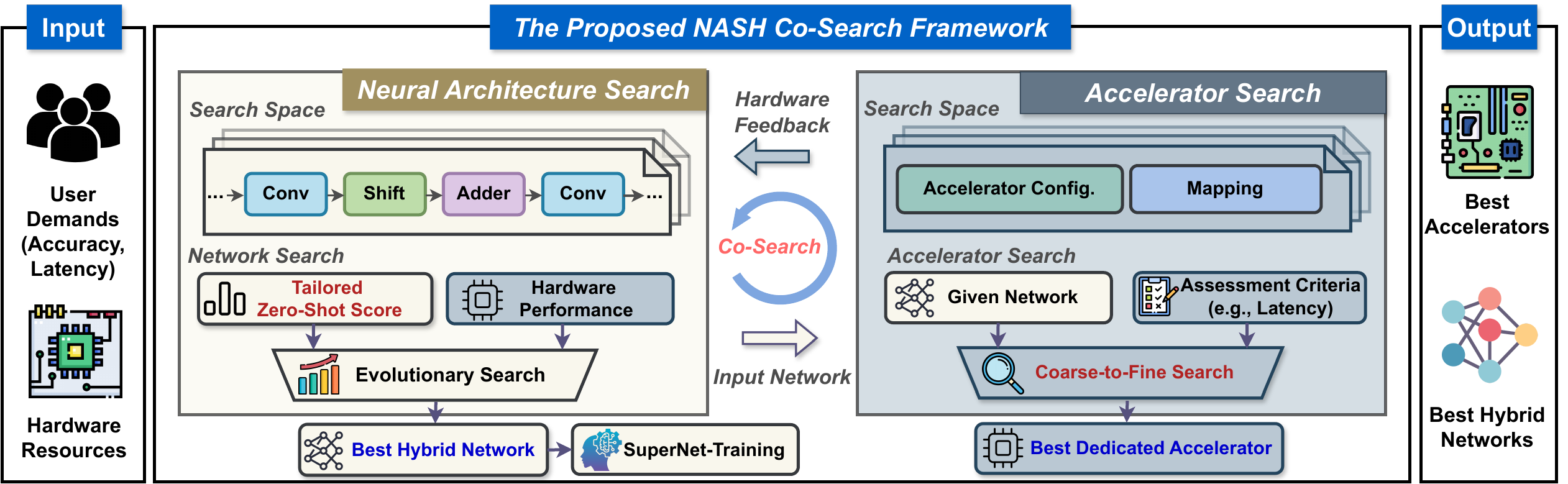}}
	\caption{The overview of our {NASH} framework, where we integrate both the neural architecture search (NAS) and coarse-to-fine accelerator search to directly obtain optimal pairing of models and accelerators. Specifically, the NAS consists of a tailored zero-shot metric to pre-identify promising multiplication-reduce hybrid models before supernet training. Besides, the accelerator search involves a novel coarse-to-fine search strategy to expedite the accelerator search process.}
    %The overview of our proposed NASA-F framework, which incorporates a dedicated one-shot supernet-based NAS engine (left) to enable the effective search for multiplication-reduced hybrid models. Additionally, NASA-F further integrates a dedicated accelerator engine (right) to fully utilize heterogeneous hardware resources on FPGAs, such as DSP slices and LUTS, for accelerating searched hybrid models with boosted hardware utilization and throughput.}
	\label{fig:overall} 
\end{figure*}

To this end, in pursuit of advancing the search efficiency, model accuracy, and hardware efficiency for multiplication-reduced hybrid models, we develop a neural architecture and accelerator co-search framework tailored to such hybrid models, and make the following contributions:
\begin{itemize}
    \item We propose \textbf{NASH} (see Fig. \ref{fig:overall}), a \textbf{N}eural architecture and \textbf{A}ccelerator co-\textbf{S}earch framework dedicated to multiplication-reduced \textbf{H}ybrid models. To the best of our knowledge, this is the first model-accelerator co-search framework for such hybrid models.
    % \item On the algorithm level, we propose a tailored \textbf{Pre-Search One-Shot NAS Engine} to identify and optimize potentially high-quality sub-networks in our hybrid supernet. Specifically, we first search for various promising sub-networks via a \textbf{zero-shot hybrid-model-oriented} search process to create a preferred sample space, and then feed these chosen sub-networks into the subsequent training phase. 
    % % To facilitate the hybrid model training, we follow NASA-F \cite{10308526} to perform \textbf{supernet slimming} before the search process starts, determining the operation type for each layer in the incoming stages.
    % Furthermore, hardware feedback from a searched accelerator (see Sec?) will be provided for each sub-network in the \textbf{zero-shot hybrid-model-oriented} search process to identify architectures with high efficiency.
     % \item On the architecture search level, we propose a tailored \textbf{Zero-Shot-Search One-Shot-Training NAS Engine} to identify and optimize potentially high-quality architectures inside our hybrid supernet. By intently optimizing selected promising architectures instead of randomly sampled ones, gradient conflicts when training different architectures can be reduced and thus a better convergence state can be achieved to boost final accuracy.  
     % Furthermore, in the search process, zero-shot metrics will provide rapid algorithm evaluation while \textbf{Coarse-to-Fine Predictor} (see Sec?) will provide hardware feedback. Therefore, sub-architectures with high efficiency can be identified.
      \shh{\item On Neural Architecture Search (NAS) level, we develop a \textbf{tailored zero-shot metric} to pre-identify promising multiplication-reduced hybrid architectures before training, thus enhancing search efficiency as well as alleviating gradient conflicts during the subsequent supernet optimization to enhance accuracy. 
     % Furthermore, in the search process, zero-shot metrics will provide rapid algorithm evaluation while \textbf{Coarse-to-Fine Predictor} (see Sec?) will provide hardware feedback. Therefore, sub-architectures with high efficiency can be identified.
     }
    
    % \item On the hardware search level, we propose a \textbf{Coarse-to-Fine Predictor} to perform hardware search and rapidly predict the optimal hardware performance for a certain network architecture. As for the hardware search process, to deal with the enormously large search space brought by our hybrid multiplication-reduced model, we leverage a coarse-to-fine search strategy to enable a fast hardware search. Different from the methods that discretely explore the search space, i.e., the evolutionary algorithm in NAAS \cite{NAAS}, we make sure the optimal result is achieved while the search cost is also largely reduced.
    \item \shh{On the accelerator search level, considering the enormous search space of the accelerator dedicated to our hybrid models, we innovatively propose a \textbf{coarse-to-fine search strategy} to significantly expedite the accelerator search process. Besides, this accelerator search can be further integrated with the above NAS process to obtain NASH, aiming to directly obtain optimal pairing of multiplication-reduced models and dedicated accelerators.}
    
    \item Extensive experimental results consistently validate our superiority. {Particularly, we offer up to $\uparrow$$0.56\%$ accuracy on Tiny-ImageNet over the prior multiplication-reduced work NASA-F, and $\uparrow$$2.14\times$ throughput as well as $\uparrow$$2.01\times$ FPS with $\uparrow$$0.25\%$ accuracy against the state-of-the-art (SOTA) multiplication-based system on CIFAR100.} 
\end{itemize}

\section{Related Works}
\label{sec:related_work}
\subsection{Multiplication-Reduced DNNs}
% To alleviate the computational burden imposed by dominant multiplications in convolutions, prior research has introduced networks that operate without explicit multiplication operations. Instead, these networks leverage hardware-friendly alternatives such as bitwise shifts and additions, thereby contributing to the progress in developing efficient models. 
\shh{To alleviate the computational burden imposed by resource-dominant multiplications involved in DNNs, prior works have developed multiplication-free models with hardware-friendly alternatives, such as bitwise shifts \cite{DeepShift} and additions \cite{AdderNet}, \cite{AdderNetv2}, \cite{AdderTransformer}, \cite{dong2023improving}, thereby contributing to the progress in developing efficient models.}
For example, DeepShift \cite{DeepShift} introduces shift layers that approximate multiplications using power-of-two equivalents, effectively substituting multiplications with bitwise shifts.
%another shift work
AdderNets \cite{AdderNet}, \cite{AdderNetv2}, \cite{AdderTransformer}, \cite{dong2023improving} presents adder layers, which utilize the $l_1$-normal distance to assess the similarity between inputs and weights, thereby trading multiplications for additions. 
% In a recent NGD-ANN \cite{dong2023improving}, instead of a CNN, an $l_2$-norm ANN is employed as the teacher model to guide $l_1$-norm ANN's training, exploiting the strong correlation between $l_1$-norm and $l_2$-norm ANNs.
Furthermore, ShiftAddNet \cite{ShiftAddNet} integrates the aforementioned shift and adder layers, resulting in multiplication-free models primarily with both bitwise shifts and additions. 
However, despite the promising hardware efficiency of these multiplication-free DNNs, they generally suffer from lower accuracy compared to their multiplication-based counterparts \cite{dong2023improving, ShiftAddNet, NASA, nasa-f}. 
% To address this issue, multiplication-reduced models have been proposed. 
\shh{To this end, multiplication-reduced models \cite{ShiftAddNet}, \cite{NASA+}, \cite{nasa-f}, \cite{NASA}, \cite{You2022ShiftAddNASHS}, \cite{you2023shiftaddvit} are highly desired to marry the benefits of both multiplication-based and multiplication-free models, aiming to enhance hardware efficiency while maintaining accuracy.
For instance, NASA \cite{NASA} and NASA-F \cite{nasa-f} integrate multiplication-based convolutions as well as multiplication-free shift and adder layers to construct hybrid search space, on top of which, they leverage search algorithms to automatically build multiplication-reduced hybrid models. }
%mark
%Moreover, ShiftAddViT \cite{you2023shiftaddvit} reparameterizes pre-trained Vision Transformer (ViT) with a mixture of multiplication primitives, thus offering multiplication-reduced ViT.}
%This observation has spurred interest in developing multiplication-reduced hybrid DNNs, aiming to combine the strengths of both approaches.

\subsection{Neural Architecture Search (NAS)} 
% Recently, Neural Architecture Search (NAS) 
% \cite{FBNet,FBNetV2,FBNetV3,BigNAS,AlphaNet,NASA-F,NASA,NASA+}, which explores numerous possibilities in the search space to identify potentially optimal architecture structures, has emerged as an effective approach to facilitate the resources-intensive process of designing efficient neural networks. 
\shh{Neural Architecture Search (NAS) 
\cite{nasa-f},
\cite{BigNAS}, \cite{AlphaNet},  \cite{NASA}, \cite{FBNet}, \cite{FBNetV2}, \cite{FBNetV3}, which aims to search for the optimal network through enormous model architectures, has emerged as an effective approach to automatically design efficient models with saved human effort and expert knowledge.}
% Among them, \emph{one-shot supernet-based NAS} \cite{FBNet,FBNetV2,FBNetV3,BigNAS,AlphaNet,NASA-F} has recently become the new state-of-the-art approach, which typically consists of two stages: supernet training and resource-constrained architecture search. In a typical supernet training process, potential sub-networks are randomly sampled and optimized， bringing most sub-networks to a collectively well-optimized status. Then resource-constrained architecture search is performed to identify the optimal network architecture under a specific resource constraint. 
\shh{Among them, \emph{one-shot supernet-based NAS} \cite{nasa-f},
\cite{BigNAS}, \cite{AlphaNet}, \cite{FBNet}, \cite{FBNetV2}, \cite{FBNetV3} has achieved remarkable results and typically consists of two stages: supernet training and resource-constrained architecture search, as shown in Fig, \ref{fig:nas_pipeline} (a). Specifically, in the supernet training process, sub-networks are generally randomly sampled and optimized. Subsequently, the resource-constrained architecture search is applied to identify the optimal network from the well-trained search space {through accuracy evaluation} while adhering to a specific resource constraint.}
 %To enhance the supernet training process, 
 %BigNAS \cite{BigNAS} applies the sandwich rule \cite{Yu2019UniversallySN} to concurrently optimize the smallest, largest (supernet), and several randomly sampled sub-networks, 
 %boosting both the lower and upper bounds of sub-architectures.
 %where the supernet is optimized with real labels and the others undergo distillation from the supernet.
 % Among these One-Shot NAS methods, AlphaNet \cite{AlphaNet} proposes utilizing a more generalized $\alpha$-divergence instead of the standard KL-divergence to facilitate supernet training. Recently, NASA-F \cite{nasa-f} successfully performed a One-Shot NAS process on multiplication-reduced models, achieving competitive accuracy while throughput is largely improved.  
 \shh{
 % Recent work AlphaNet \cite{AlphaNet} proposes a generalized $\alpha$-divergence to trade the standard KL-divergence, thus facilitating supernet training.
 For isntance, NASA-F \cite{nasa-f} develops a tailored one-shot NAS engine for multiplication-reduced models, aiming to maintain accuracy while enhancing hardware efficiency.}

% Recently, some frontier works have explored the possibility of combining one-shot nas with zero-shot nas. Zero-shot nas can perform architecture searches without real data and by directly analyzing networks' characteristics, thus no time-consuming evaluating is needed, allowing a much more swift search process. For instance,
%  PreNAS \cite{wang2023prenas} swaps the order of training and search in classical one-shot nas, searching for various promising sub-networks in advance by a zero-shot metric, e.g., SNIP \cite{lee2018snip}, so that training resources can be concentrated in the later training phase to boost performance and alleviate the search overhead as well. However, the success of combining zero-shot and one-shot nas on multiplication-based models is yet to be transplanted to multiplication-reduced models. 
\shh{{%Recently, some frontier works have advanced one-shot NAS by incorporating zero-shot metrics. As depicted in Fig. \ref{fig:nas_pipeline} (b), these metrics are applied to identify potentially proficient architectures before network training, aiming to alleviate gradient conflict during the subsequent supernet training to boost accuracy \cite{wang2023prenas}. 
Recently, some frontier works \cite{wang2023prenas,lee2018snip,nn_degree,zen-nas,grasp} have proposed zero-shot metrics, aiming to pre-identify potentially promising architectures without model training.}
%其实要说的话prenas没有propose zero-shot metric， 它是直接用了别人的
%Particularly, PreNAS \cite{wang2023prenas} swaps
{Motivated by this,
PreNAS \cite{wang2023prenas} advocates swaping} the order of supernet training and search process in the standard one-shot NAS. Specifically, as depicted in Fig. \ref{fig:nas_pipeline} (b), it employs a zero-shot metric (i.e., SNIP \cite{lee2018snip}) to predict promising sub-networks, then allocates training resources to optimize the selected ones instead of randomly sampled ones to facilitate supernet training. However, zero-shot metrics tailored for multiplication-reduced hybrid models are still under-explored. }

\subsection{Accelerator Search}

\shh{Considering the enormous design space within a dedicated accelerator, including (i) the hardware configurations specifying the hardware resource consumption in terms of buffers and computational resources and (ii) the mapping method (also dubbed dataflow) that indicates how computations are scheduled onto the accelerator, it is non-trivial to manually craft an optimal accelerator, which demands substantial expertise and iterative trials \cite{NAAS,zhang2020dna,hong2023dosa}.
To solve this limitation, accelerator search methodologies \cite{NAAS,zhang2020dna,hong2023dosa} have been developed to automatically identify both the optimal hardware configuration and mapping method, thus enhancing hardware efficiency.
%mark
%Particularly, DOSA \cite{hong2023dosa} introduces a fast gradient-based search algorithm to expedite the accelerator search process.
Additionally, the accelerator exploration can be further incorporated with the neural architecture search to directly obtain optimal pairing of models and accelerators \cite{NAAS,zhang2020dna}. For instance, NAAS \cite{NAAS} formulates the network and accelerator co-search as a multi-loop process and employs an evolutionary algorithm to facilitate this complex procedure.
Despite the effectiveness of the above works, they are exclusively tailored to designing accelerators for homogeneous networks that are characterized by merely multiplication-based operations.
In light of the heterogeneous layers in our desired multiplication-reduced hybrid models, which include both multiplication-based and multiplication-free layers, the search space within the accelerator dedicated to such hybrid models becomes more intricate \cite{NASA, NASA+}, calling for more effective search solutions.}
\section{The Neural Architecture Search} \label{sec:neural architecture search}
%\shh{In this section, we first introduce our hybrid search space that integrates both multiplication-based convolutions and multiplication-free operations}; then the tailored zero-shot metric used in our proposed network search strategy to accurately predict the model accuracy with little resource consumption; finally, the details of the network search strategy which further boosts the model accuracy while alleviating the search cost.
% In this section, we present the neural architecture search strategy without accelerator search included. Specifically, {we first introduce our hybrid search space that integrates both multiplication-based convolutions and multiplication-free operations}; then the tailored zero-shot metric used in our proposed neural architecture search strategy to accurately predict the model accuracy with little resource consumption; finally, the details of the neural architecture search strategy which further boosts the model accuracy while alleviating the search cost.
\shh{In this section, we first introduce our hybrid search space that integrates both multiplication-based convolutions and multiplication-free operations; Then, Sec. \ref{sec:nas} illustrates our zero-shot search strategy that is equipped with a tailored zero-shot metric to pre-identify promising sub-networks within our hybrid search space before network training; Finally, Sec. \ref{sec:training} details our preference-biased supernet training, which concentrates training resources on these selected sub-networks, aiming to alleviate gradient conflict and boost accuracy.}

\subsection{The Hybrid Search Space}
\label{sec:search_space}

\textbf{Fundamental Operations.} 
% We simultaneously include multiplication-based \textbf{\emph{convolutions}} and multiplication-free \textbf{\emph{shift}} and \textbf{\emph{adder layers}} into our hybrid search space to enable searching for multiplication-reduced hybrid models. Next, we will give introductions to shift and adder layers.
\shh{To effectively search for desired multiplication-reduced models, we unify multiplication-based \textbf{\emph{convolutions}} and multiplication-free \textbf{\emph{shift}} and \textbf{\emph{adder layers}} to construct our hybrid search space following \cite{nasa-f}. Next, we will illustrate shift and adder layers, respectively.}
\begin{itemize}
    \item \textbf{Shift layers.} 
    % Multiplication can be replaced with a more hardware-friendly bitwise shift operation if one of the multiplication factors is a power of two. As demonstrated in \cite{DeepShift}, the weights in shift layers are exclusively in power-of-two formats, resulting in enhanced hardware efficiency.
    % To update the discrete power-of-two weights, a full-precision counterpart $W_{\text{Conv}}$ is created and optimized during backward training. The power-of-two weights are then generated by quantizing the full-precision $W_{\text{Conv}}$. In this context, we denote the inputs and weights of shift layers as $X$ and $W_{\text{Shift}}$, respectively. The processing of shift layers can be described as follows: 
    \shh{To enhance hardware efficiency, as outlined in Eq. (\ref{eq:shift}), DeepShift \cite{DeepShift} advocates the utilization of shift layers, where the weights $W_{\text{Shift}}$ are derived by quantizing the vanilla convolutional weights $W_{\text{Conv}}$ to their power-of-two equivalents following Eq. (\ref{eq:shift-q}). By doing this, the costly multiplications in convolutions can be effectively substituted by hardware-efficient bitwise shifts.}
    \begin{equation}
		Y = \sum X^\mathrm{T}*W_{\text{Shift}}, \label{eq:shift}
    \end{equation}
    
    \begin{equation}
    % \small
	\begin{aligned}
		W_{\text{Shift}}=\hat{s}*2^{\hat{p}}, \ \text{where}& \\ \hat{s}=\text{sign}(W_\text{Conv}), \ \hat{p}=\text{round}(\text{log}&_{2}|W_\text{Conv}|).
		\end{aligned} 
  \label{eq:shift-q}
    \end{equation}
    
    \item \textbf{Adder layers.} 
    % AdderNet \cite{AdderNet} introduced adder layers that compute the $l_1$-norm distance to measure the relevance between activations and weights, eliminating conventional multiplication-based convolution layers. Specifically, we represent the inputs and weights of adder layers as $X$ and $W_{\text{adder}}$ respectively. The computation within adder layers can be described as follows:
    \shh{As an alternative, as formulated in Eq. (\ref{eq:adder}), AdderNet \cite{AdderNet} builds adder layers that employ the $l_1$-norm distance to measure the relevance between activations $X$ and weights $W_{\text{adder}}$, thus trading multiplications with efficient additions.}
    \begin{equation}
		Y =\sum -\left\lvert X- W_{\text{adder}}\right\rvert. 
		\label{eq:adder}
    \end{equation}
\end{itemize}

\begin{table}[]
\centering
\caption{Choices of the channel number $C$, expansion ratio $E$, kernel size $K$, operation type $T$, and the block number $N$ in our hybrid search space. \{\texttt{C}, \texttt{S}, \texttt{A}\} denote \{convolutions, shift layer, adder layer\}. \texttt{MBPool} represents the efficient last stage \cite{mbv3}}
\resizebox{0.95\linewidth}{!}{
\setlength{\tabcolsep}{0.35em}
\begin{tabular}{c|ccccc}
\Xhline{2.5\arrayrulewidth}
\textbf{Block}      & \textbf{C}                 & \textbf{E}              & \textbf{K}    & \textbf{T}      & \textbf{N}                   \\ \hline \hline
\textbf{First Conv} & {\{16, 24\}}                & {-}                     & {1}           & \texttt{C}             & {-}                 \\
\textbf{Stage 1}    & {\{16, 24\}}               & {\{1\}}             & {\{3, 5\}}    & {\{\texttt{C}, \texttt{S}, \texttt{A}\}}         & {\{1, 2\}}   \\
\textbf{Stage 2}    & {\{24, 32\}}               & {\{4, 5, 6\}}           & {\{3, 5\}}    & {\{\texttt{C}, \texttt{S}, \texttt{A}\}}   & {\{3, 4, 5\}}   \\
\textbf{Stage 3}    & {\{32, 40\}}               & {\{4, 5, 6\}}        & {\{3, 5\}}    & {\{\texttt{C}, \texttt{S}, \texttt{A}\}}   & {\{3, 4, 5, 6\}}   \\
\textbf{Stage 4}    & {\{64, 72\}}               & {\{4, 5, 6\}}        & {\{3, 5\}}    & {\{\texttt{C}, \texttt{S}, \texttt{A}\}}   & {\{3, 4, 5, 6\}}   \\
\textbf{Stage 5}    & {\{112, 120, 128\}}        & {\{4, 5, 6\}}  & {\{3, 5\}}    & {\{\texttt{C}, \texttt{S}, \texttt{A}\}}   & {\{3, 4, 5, 6, 7, 8\}}   \\
\textbf{Stage 6}    & {\{192, 200, 208, 216\}}   & {\{6\}}  & {\{3, 5\}}    & {\{\texttt{C}, \texttt{S}, \texttt{A}\}}         & {\{3, 4, 5, 6, 7, 8\}}   \\
\textbf{Stage 7}    & {\{216, 224\}}             & {\{6\}}              & {\{3, 5\}}    & {\{\texttt{C}, \texttt{S}, \texttt{A}\}}         & {\{1, 2\}}   \\
\textbf{MBPool}  & {\{1792, 1984\}}           & -                       & {1}           & \texttt{C}             & {-}           \\ \Xhline{2.5\arrayrulewidth}
\end{tabular}} \label{tab:search_space} 
\end{table}

\textbf{Supernet.} 
% Given the above basic operations, we present our hybrid search space in Table \ref{tab:search_space}. 
% On top of it, we build our supernet by incorporating all candidate sub-networks from the pre-defined search space via a weight-sharing strategy. 
\shh{By incorporating the above fundamental operations, we follow \cite{nasa-f} to build our hybrid supernet. Specifically, As listed in Table \ref{tab:search_space}, the supernet mainly consists of seven stages, and each stage is composed of several inverted residual bottleneck (IRB) blocks \cite{mobilenetv2} that comprise two point-wise (PW) layers separated by one depth-wise (DW) layer. There are five searchable parameters in each stage to specify a sub-network from the supernet: the output channel number of blocks/layers $C$, the channel expanding ratio of IRBs $E$, the kernel size of DW layers inside IRBs $K$, the layer type $T$, and the number of IRBs within a stage $N$.} 

\subsection{Zero-Shot Search}
\label{sec:nas}
\subsubsection{The Tailored Zero-shot Metric}
% To quickly identify high-quality sub-networks within our pre-defined hybrid search space, we leverage the assistance of zero-shot metrics. 
\shh{To alleviate gradient conflict during supernet training and boost accuracy,
zero-shot metrics are highly desired to pre-identify high-quality sub-networks within our pre-defined hybrid search space before network optimization.}

\textbf{The Observation and Challenge.}
\shh{Zero-shot metrics tailored for multiplication-based models \cite{lee2018snip,gradnorm,grasp,fisher,jacob} have been widely explored and achieved remarkable success. However, we have observed non-negligible performance degradation when directly applying them to our multiplication-reduced models. Specifically, as demonstrated in Table \ref{tab:zero-shot metrics}, the Kendall Tau Coefficient \cite{kendall1938new}, a widely-used similarity measurement metric, between existing popular zero-shot metrics \cite{lee2018snip}, \cite{grasp}, \cite{synflow}, \cite{fisher}, \cite{gradnorm}, \cite{jacob}, \cite{xiong2020LINEARREGIONnumber} and model accuracy is obviously higher in multiplication-based models than in our multiplication-reduced hybrid models. This underscores the necessity for customized solutions for
our hybrid models, which is yet under-explored. }

\begin{table}[]
\centering
\caption{The Kendall Tau Coefficient between zero-shot metric scores and real model accuracy. 
% We randomly sample 300 sub-networks from a well-optimized supernet in NASA-F \cite{nasa-f} and evaluate their accuracy as the real model accuracy. 
$T$ and $E$ are the abbreviations for trainability and expressivity}
\resizebox{0.9\linewidth}{!}{
\setlength{\tabcolsep}{0.35em}
\begin{tabular}{c|ccc}
\Xhline{2.5\arrayrulewidth}
\textbf{Metrics}   & \multicolumn{1}{l}{\textbf{Mult.-based}} & \multicolumn{1}{l}{\textbf{Mult.-reduced}} & \textbf{Class} \\ \hline \hline
SNIP \cite{lee2018snip}               & 0.40                                               & -0.01                                                 & T \\ \hline
Jacobian Covariant \cite{jacob}              & 0.43                                               & 0.06                                                 & T \\ \hline
Grad Norm \cite{gradnorm}          & 0.36                                               & -0.09                                                 & T \\ \hline
Synflow \cite{synflow}            & 0.47                                              & 0.06                                                 & T \\ \hline
Grasp \cite{grasp}              & 0.35                                               & 0.03                                                 & T \\ \hline
Fisher \cite{fisher} & 0.45                                               & -0.05                                                 & T \\ \hline
\rowcolor{orange!15} NN-Degree \cite{nn_degree} & 0.43                                               & 0.32                                                 & T \\ \hline
Linear Region Number \cite{xiong2020LINEARREGIONnumber}      & 0.32                                               & 0.19                                                 & E \\ \hline
\rowcolor{orange!15} Zen-Score \cite{zen-nas,zenDet}      & 0.40                                               & 0.33                                                 & E \\ \hline
\rowcolor{dark-green!15} \textbf{Ours}       & -                                               & \textbf{0.47}                                                 & \textbf{T\&E} \\ \Xhline{2.5\arrayrulewidth}
\end{tabular}} \label{tab:zero-shot metrics} 
\end{table}

\textbf{Our Proposed Solution.}
% Existing zero-shot metrics commonly assess a network's performance based on its expressivity or trainability \cite{zeroshotsurvey, TE-NAS}. To compensate for the accuracy degradation introduced above, our tailored zero-shot metric, inspired by the approach of TE-NAS \cite{TE-NAS}, concurrently assesses a network's expressivity and trainability by combining two zero-shot metrics. However, determining which zero-shot metric can accurately predict a multiplication-reduced model's expressivity or trainability is still an area under exploration.
\shh{Existing zero-shot metrics typically assess networks' performance based solely on trainability or expressivity \cite{zeroshotsurvey, TE-NAS}, resulting in biased measurements. To enable more accurate assessments, inspired by TE-NAS \cite{TE-NAS}, we assess both the expressivity and trainability of models by integrating distinct zero-shot metrics, where the pivotal challenge lies in identifying effective zero-shot metrics tailored for our multiplication-reduced models.}

\paragraph{\textbf{Trainability}}
\shh{Models with high trainability can be effectively optimized via gradient descent, thus demonstrating high accuracy. 
To distinguish a powerful zero-shot metric for evaluating the trainability of our multiplication-reduced models, 
we first intuitively explore the widely-adopted gradient-based methods, including SNIP \cite{lee2018snip}, Jacob covariance \cite{jacob}, Grad Norm \cite{gradnorm}, Synflow \cite{synflow}, Grasp \cite{grasp}, and Fisher \cite{fisher}.
However, as depicted in Table \ref{tab:zero-shot metrics}, they all suffer from severe performance drops.
% due to their reliance on disrupted weights and corresponding gradients. 
To gain deeper insights into this degradation, we take the representative gradient-based metric SNIP \cite{lee2018snip} as an illustrative example. Specifically, SNIP assesses model trainability by measuring the importance of its parameters in both the forward and backward processes and can be defined as follows: }
\begin{equation}
\mathrm{SNIP} = \sum_{i=1}^{N}\left|\left\langle\boldsymbol{\theta}_{i}, \nabla_{\boldsymbol{\theta}_{i}} \mathcal{L}\right\rangle\right|,
\end{equation}
where $\langle \cdot \rangle$ donates the inner product, $N$, ${\theta}_{i}$ and $\mathcal{L}$ are the number of layers, the parameter vector of the $i$-th layer within the given network, and the loss value, respectively. 
% However, in our multiplication-reduced hybrid models, weights and gradients of multiplication-free layers behave differently from those of convolution layers. 
% For example, the value magnitude of weights and gradients in adder layers are significantly larger than those in convolution layers \cite{AdderNet}, whereas weights are discrete and gradients are biased in shift layers \cite{DeepShift}. In this manner, multiplication-reduced models involve multiple heterogeneous weights and gradients that coexist and interact, blocking gradient-based methods designed for homogeneous multiplication-based models to capture useful information.
\shh{Unfortunately, in our multiplication-reduced hybrid models that includes both multiplication-free shift and adder layers alongside multiplication-based convolutions, occasions are more complicated. 
For example, weights are discrete and gradients are biased in shift layers \cite{DeepShift}. Moreover, the value magnitude of both weights and gradients in adder layers are significantly larger than those in convolutions \cite{AdderNet}. Therefore, due to the distinct behaviors of heterogeneous layers concerning weights and gradients, existing gradient-based zero-shot metrics designed for homogeneous models are inherently inapplicable to our hybrid models, yielding performance degradation.}
%As a result, gradient-based methods fail to capture useful information for our multiplication-reduced models.

\begin{figure}[t]
    \centerline{\includegraphics[width=0.93\linewidth]{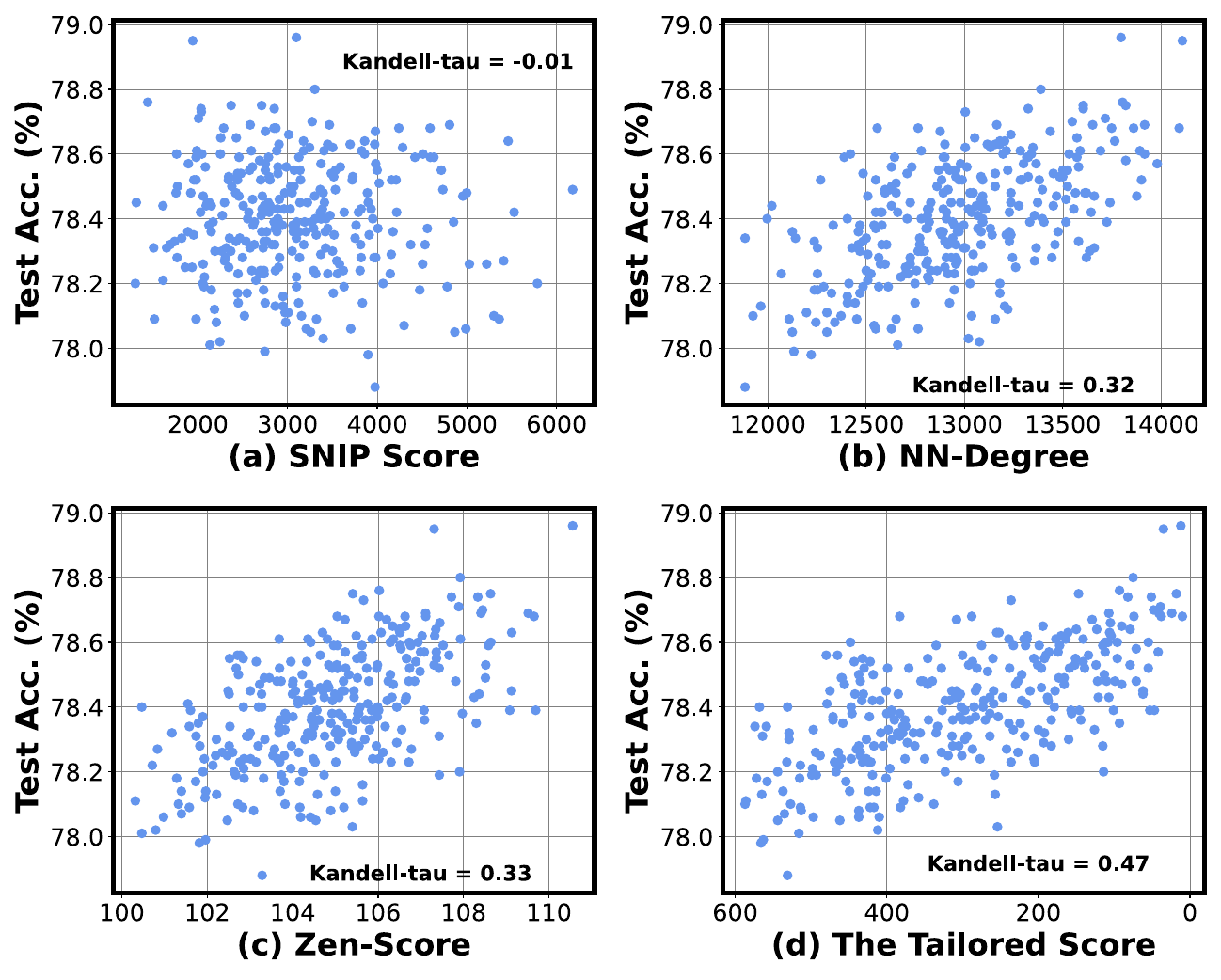}}
    \caption{Correlations between model accuracy and zero-shot metrics, including (a) SNIP, (b) NN-Degree, (c) Zen-Score, and (d) our tailored zero-shot metric, when measured on multiplication-reduced hybrid models.
    % , (c) SOTA DSP packing strategy for additions \cite{Xilinx-add} that accommodates \textbf{four} $8$-bit additions using one DSP slice, and (d) the DSP-LUT co-packing strategy \cite{Zhang2022WSQAdderNetEW} that allows \textbf{eight} $8$-bit additions using integrated adders of DSP slices with the help of additional LUTs.
    }
	\label{fig:metric_accuracy}
\end{figure}

% This ineffectiveness of gradient-based methods prompts us to an alternative approach that assesses a neural network’s trainability by analyzing its connectivity pattern (e.g., the topology of concatenation-type skip connections) \cite{nn_mass,nngp,nn_degree}. 
\shh{To overcome this issue, we redirect our attention to the connectivity-based methods \cite{nn_mass,nngp,nn_degree}, which evaluate trainability through the analysis of models' connectivity patterns (e.g., the topology of concatenation-type skip connections \cite{nn_degree}).}
%To achieve this, we recognize that some prior works \cite{nn_mass,nngp,nn_degree} propose assessing a neural network’s trainability by analyzing its connectivity pattern (e.g., the topology of concatenation-type skip connections) and have achieved remarkable success. 
% We advocate these connectivity-based methods are more appropriate for hybrid models as they do not require to consider the intricate details of weights and gradients within multiplication-reduced models and offer faster evaluation since no gradient operation is involved.
\shh{Particularly, we select NN-Degree \cite{nn_degree}, which is the SOTA one and can be formulated as:}
%\begin{equation} \label{eq:nn-degree}
    %\text{NN-Degree}=\sum_{\text{each block } c}\left(w_{c}+\frac{\text{\# Residual connections}}{\text {\# Total input channels}}\right),
%\end{equation}
\begin{equation} \label{eq:nn-degree}
    \text{NN-Degree}=\sum_{i=1}^{B}\left(\frac{\sum_{j=1}^{N_i}C_{i,j}^{O}}{N_i}+\frac{C_{i}^{R}}{\sum_{j=1}^{N_i}C_{i,j}^{I}}\right),
\end{equation}
% where $w_c$ is the average width value of a network block $c$. 
% As demonstrated in Table. \ref{tab:zero-shot metrics} and Fig. \ref{fig:metric_accuracy}, NN-Degree maintains reasonable accuracy on hybrid models.
\shh{where $B$ denotes the total block number in a given network and $N_i$ is the layer number of the $i$-th block. $C_{i,j}^{O}$ and $C_{i,j}^{I}$ are the output and input channel number of the $j$-th layer in the $i$-th block, and $C_{i}^{R}$ is the residual connection channel number of the $i$-th block.
%where $w_c$ is the average width value of a network block $c$. 
As verified in Table \ref{tab:zero-shot metrics} and Fig. \ref{fig:metric_accuracy}, while NN-Degree does not exhibit the best predictive performance on multiplication-based models, it outperforms other gradient-based methods on our hybrid models. {Besides, it enables faster evaluation as no gradient computation is involved.}}

\paragraph{\textbf{Expressivity}}
% Expressivity refers to the expressive capability of models, yet the number of existing zero-shot metrics evaluating expressivity is still limited. 
% In order to identify the most effective metric for the expressivity of our hybrid models, we conducted experiments on two of the most well-known ones: Linear Regions Number \cite{xiong2020LINEARREGIONnumber} and Zen-score \cite{zen-nas,zenDet}.
\shh{Expressivity refers to the expressive capability of models. 
To identify the most effective zero-shot metric for the expressivity assessment of our hybrid models, we conducted experiments on two of the most well-known expressivity-based metrics: Linear Regions Number \cite{xiong2020LINEARREGIONnumber} and Zen-Score \cite{zen-nas,zenDet}.}
% Among them, Linear Regions Number \cite{xiong2020LINEARREGIONnumber} and Zen-score \cite{zen-nas,zenDet} stand out as two well-known metrics. 
%Specifically, Linear Region Number assesses the expressivity of a network by counting the number of unique linear regions within its input space, while Zen-Score leverages the gradient of input $x$ to analyze expressivity. 
% Specifically, Linear Region Number assesses the expressivity of a network by counting the number of unique linear regions within its input space. Zen-Score measures the Gaussian complexity of a network to evaluate its expressivity and is formulated as follows: 
\shh{Specifically, Linear Region Number assesses the expressivity of models by counting the number of unique linear regions within their input space \cite{xiong2020LINEARREGIONnumber}. Zen-Score measures models' Gaussian complexity to evaluate their expressivity \cite{zen-nas,zenDet} and can be formulated as follows: }
\begin{equation} \label{eq:zenscore}
\begin{array}{c}
\text{Zen-Score} = \log \mathbb{E}_{\boldsymbol{x}, \boldsymbol{\epsilon}}\left(\left\|f_{e}(\boldsymbol{x})-f_{e}(\boldsymbol{x}+\alpha \boldsymbol{\epsilon})\right\|_{F}\right) \\[0.5em]
+ \sum_{k, i} \log \left(\sqrt{\frac{\sum_{j} \sigma_{i j}^{k}}{C^{O}_{i}}}\right), 
%\\ \boldsymbol{x} \sim \mathcal{N}(0, \boldsymbol{I}),
\end{array} 
\end{equation}
\shh{where $x$ is a sampled Gaussian random vector, $\epsilon$ is a small input perturbation, $\|\cdot\|_{F}$ is the Frobenius norm, $\alpha$ is a tunable hyper-parameter, $C_{i}^{O}$ is the number of output channels of the $i^{th}$ layer, and $\sigma_{i j}^{k}$ is the variance of the $k^{th}$ sample in an input batch data for the $i^{th}$ layer’s $j^{th}$ channel.
As validated in Table \ref{tab:zero-shot metrics}, Zen-Score shows superior estimation performance over Linear Region Number on both multiplication-based models and multiplication-reduced hybrid models.
% and we attribute this to the following reasons, (i) the inherent superiority of Zen-Score, e.g., higher accuracy on multiplication-based models \cite{zen-nas}; (ii) Zen-Score can maintain its accuracy in our hybrid models since no weight gradient is involved in Eq. (\ref{eq:zenscore}).
%(gradient of input reflects the influence of multiplication-free operations whereas Linear Region Number fails to distinguish between linear regions of different operations).
Additionally, it also exhibits better computational efficiency as counting the unique linear region number of large models involved in Linear Region Number has been proved to be very time-consuming \cite{zen-nas}.
Therefore, we select Zen-Score to assess the expressivity of our multiplication-reduced models. }
%Precisely, Zen-Score is formulated as follows:
%\begin{equation} \label{eq:zenscore}
%\begin{array}{c}
%\text{Zen-Score} = \log \mathbb{E}_{\boldsymbol{x}, \boldsymbol{\epsilon}}\left(\left\|f_{e}(\boldsymbol{x})-f_{e}(\boldsymbol{x}+\alpha \boldsymbol{\epsilon})\right\|_{F}\right)+\\ \sum_{k, i} \log \left(\sqrt{\frac{\sum_{j} \sigma_{i j}^{k}}{C h_{i}}}\right), \\
%\boldsymbol{x} \sim \mathcal{N}(0, \boldsymbol{I}),
%\end{array} 
%\end{equation}
%where $\|\cdot\|_{F}$ is the Frobenius norm, $\alpha$ is a tunable hyper-parameter, $Ch_{i}$ is the number of channels of the $i$-th layer and $\sigma_{i j}^{k}$ is the variance of the $i$-th layer’s $j$–th channels for the $k$-th samples in an input batch data.

% On top of the above two carefully chosen metrics, we follow TE-NAS \cite{TE-NAS} to
% On top of the above two carefully chosen metrics, 
% As depicted in Table. \ref{tab:zero-shot metrics}, although the degradation of the aforementioned NN-Degree and Zen-score is relatively small, to more accurately assess hybrid networks, we need to combine these two metrics effectively.
% % can accurately predict the network's expressivity or trainability, 
% However, as illustrated in Fig. \ref{fig:metric_accuracy}, the scores of these two metrics exhibit significant differences in magnitudes, directly summing them might result in one metric numerically overwhelming the other. 
% To address this, we follow TE-NAS \cite{TE-NAS} to add the relative rankings of these two scores, alleviating the impact of their varying magnitudes. Precisely, given a group of networks $\mathcal{N}$, the score of our tailored zero-shot metric for a specific network $\alpha$ is defined as follows:
\paragraph{\textbf{Overall}}
\shh{To enable a more accurate assessment of our hybrid networks, we further integrate the aforementioned two selected zero-shot metrics.}
%However, as illustrated in Fig. \ref{fig:metric_accuracy}, the scores of these two metrics exhibit significant differences in magnitudes, directly summing them will result in one metric numerically overwhelming the other. 
%To address this, we follow TE-NAS \cite{TE-NAS} to add the relative rankings of these two scores, alleviating the impact of their varying magnitudes. Precisely, given a group of networks $\mathcal{N}$, the score of our tailored zero-shot metric for a specific network $\alpha$ is defined as follows:}
% Specifically, considering the significant difference in score magnitudes between these two metrics, as illustrated in Fig. \ref{fig:metric_accuracy}, we follow TE-NAS \cite{TE-NAS} to add the relative rankings of these two scores. Precisely, given a group of networks $\mathcal{N}$, the score of our tailored zero-shot metric for a specific network $\alpha$ is defined as follows:
\shh{Specifically, given the significant difference in score \textit{magnitudes} between these two metrics, as illustrated in Figs. \ref{fig:metric_accuracy} (b) and \ref{fig:metric_accuracy} (c), we add the relative \textit{rankings} \cite{TE-NAS} instead of magnitudes of these two scores. Formally, given a group of networks $\mathcal{N}$, the score of our tailored zero-shot metric for a specific network $\alpha$ is defined as follows:}
\begin{equation} \label{eq:score}
\begin{split}
    \text{Score}(\alpha, \mathcal{N}) = \text{rank}(\text{Zen-Score}(\alpha), \text{Zen-Score}(\mathcal{N})) + \\ \text{rank}(\text{NN-Degree}(\alpha), \text{NN-Degree}(\mathcal{N})),
\end{split}
\end{equation}
% where the first term returns the relative ranking number of the Zen-Score of network $\alpha$ among the Zen-Scores of the network group $\mathcal{N}$, e.g. if $\alpha$ has the highest Zen-Score among $\mathcal{N}$, it returns $0$. The second term is the same case but for NN-Degree.
% % $\text{rank}$ indicates the relative ranking position of a score within scores of the network group $\mathcal{N}$. Notably, a lower rank score indicates a higher level of performance prediction. 
% The effectiveness of our tailored zero-shot metric is depicted in Fig. \ref{fig:metric_accuracy}, with the Kendall-tau correlation as 0.46. 
\shh{where the first term computes the relative ranking of the Zen-Score of network $\alpha$ within the Zen-Scores of the network group $\mathcal{N}$. 
For instance, if $\alpha$ exhibits the highest Zen-Score among $\mathcal{N}$, the term yields a value of $0$. 
%mark
%The subsequent term operates similarly, but it applies to NN-Degree instead.
Table \ref{tab:zero-shot metrics} and Fig. \ref{fig:metric_accuracy} verify the effectiveness of our tailored zero-shot metric, which showcases the highest Kendall-Tau Correlation.} 
\shh{It is noteworthy that our proposed metric also contributes to enhanced search efficiency, owing to the swift computational speed of both NN-Degree and Zen-Score. 
% For example, the accuracy evaluation of a single hybrid model derived from our supernet on average consumes $30$ seconds, whereas computing our tailored zero-shot metric merely requires less than $2$ seconds. 
For example, the assessment of accuracy for an individual hybrid model derived from our supernet takes an average of $30$ seconds, whereas the computation of our tailored zero-shot metric requires less than $2$ seconds, which is over $\mathbf{15\times}$ faster when tested on CIFAR100 and profiled on an NVIDIA GeForce RTX 2080Ti.}
\subsubsection{Neural Architecture Search}

On top of the tailored zero-shot metric, we leverage the evolutionary algorithm (i.e., the genetic algorithm) \cite{RegularizedEF} to expedite the pre-identification of promising sub-networks, aiming to alleviate gradient conflicts during the subsequent supernet training to boost accuracy. In detail, we first (i) randomly sample a population of sub-networks $\mathcal{A}$ from our pre-defined hybrid supernet (see Sec. \ref{sec:search_space} and Table \ref{tab:search_space}); Then, (ii) we expand the population $\mathcal{A}$ by crossover and mutation;
Subsequently, (iii) we update $\mathcal{A}$ by ranking candidates based on the score of our tailored zero-shot metric following Eq. (\ref{eq:score}) and retaining only the top-k ones $\widetilde{\mathcal{A}}$, subject
to given hardware constraints.
Note that to enable fast and accurate estimations, we follow \cite{NASA, nasa-f} to build a cycle-accurate chip simulator on top of our dedicated accelerator (which will be introduced in Sec. \ref{sec:accelerator search}) to measure hardware performance.
% \begin{equation}
%     \label{eq:A_c}
%     \mathcal{A}_{[c]} =  \{\alpha \mid \alpha \in \mathcal{A} \ \textbf{s.t.} \ r(\alpha) < c \},
% \end{equation}
% \begin{equation}
%     C =[c_1, c_2, c_3, \cdots],
% \end{equation}
% where $\mathcal{A}$ is the total sub-networks in existing expanded population, $\mathcal{A}_{[c]}$ is the group satisfying $c$, and $r(\alpha)$ represents the hardware performance for a specific sub-network $\alpha$, notably, it can be obtained by any hardware performance measurement methods, including FLOPs, MACs, or the hardware feedback from our searched accelerator which will be introduced in Sec. \ref{sec:accelerator search};
% (iv) For each group, keep the top $N$ scoring sub-networks to become the new population:
% \begin{equation} 
%     \widetilde{\mathcal{A}}_{[c]}=\{\alpha \mid \alpha \in \mathcal{A}_{[c]} \ \textbf{s.t.} \ \text{Score}(\alpha,\mathcal{A}_{[c]}) \in \text{Top-N}\},
% \end{equation}
% \begin{equation}
% \begin{split}
%     \widetilde{\mathcal{A}}=\{\widetilde{\mathcal{A}}_{[c]} \mid c \in C\}, %\\ \text{where} \ C =(c_1, c_2, c_3, \cdots),
% \end{split}
% \end{equation}
% where $\widetilde{\mathcal{A}}_{[c]}$ is the top-N sub-networks in $\mathcal{A}_{[c]}$ and $\widetilde{\mathcal{A}}$ is the new population. 
Finally, steps (ii) to (iv) are iterated until the pre-determined iteration number is reached. The algorithm pipeline is outlined in Alg. \ref{alg:NAAS-H} and will be detailed in Sec. \ref{sec:hw_naas}.

\subsection{Preference-Biased Supernet Training}
\label{sec:training}
% For the sequential supernet training, a one-shot supernet training process will be performed, consistently sampling sub-networks from $\widetilde{\mathcal{A}}$ and optimizing them. 
% As a result, training resources are more concentrated on these selected sub-networks compared to the vanilla one-shot supernet training and thus higher model accuracy is achievable. 
% Moreover, to facilitate the supernet training, similar to NASA-F \cite{nasa-f}, the sandwich-rule-guided architecture sampling method \cite{BigNAS} and $\alpha$-divergence-based knowledge distillation \cite{AlphaNet} are leveraged.
% Precisely, the training process can be defined as:
\shh{After identifying promising hybrid sub-networks through our zero-shot search, training resources can be concentrated on these selected sub-networks via preference-biased supernet training, aiming to boost accuracy. 
To facilitate this process, we leverage the SOTA supernet training strategy in \cite{nasa-f, AlphaNet}, which includes a sandwich-rule-guided architecture sampling \cite{BigNAS} and an $\alpha$-divergence-based knowledge distillation \cite{AlphaNet}.
Precisely, the training process can be defined as:}
\begin{equation}
\label{eq:training}
	\mathop{min}\limits_{\theta}\mathbb{E}_{\alpha_r \sim \widetilde{\mathcal{A}}}
	\{\mathcal{L}_{\text{CE}}(\theta,\alpha_b) + \gamma [ \mathcal{L}_{{\text{KD}}}(\theta,\alpha_s)+ \sum_{i=1}^{M}\mathcal{L}_{\text{KD}}(\theta,\alpha_r^i)]\},
\end{equation}
% where $\theta$ is the supernet weights, ${L}_{\text{CE}}$ denote the cross entropy loss function while the sandwich rule is applied to simultaneously optimize the smallest sub-network $\alpha_s$, the biggest sub-network $\alpha_b$, and some random networks $\alpha_r$ from our pre-defined sub-networks pool $\widetilde{\mathcal{A}}$. Moreover, the $\alpha$-divergence-based knowledge distillation process leverages the soft logits from $\alpha_b$ to optimize $\alpha_r$ and $\alpha_s$ through $\alpha$-divergence which defines $\mathcal{L}_{{\text{KD}}}$ as follows: 
\shh{where $\theta$ is the supernet weights, $\mathcal{L}_{\text{CE}}$ denote the cross entropy loss, and $\gamma$ is the loss coefficient. Besides, the sandwich rule is applied to simultaneously optimize the smallest sub-network $\alpha_s$, the biggest sub-network $\alpha_b$, and $M$ random networks $\alpha_r$ from our pre-defined sub-networks pool $\widetilde{\mathcal{A}}$, thus pushing forward both the performance lower bound ($\alpha_s$) and upper bound ($\alpha_b$) of $\widetilde{\mathcal{A}}$. Additionally, $\mathcal{L}_{{\text{KD}}}$ represent the $\alpha$-divergence-based knowledge distillation, which leverages the soft logits from $\alpha_b$ to optimize $\alpha_r$ and $\alpha_s$ through $\alpha$-divergence, aiming to alleviate the issue of under-estimation and
over-estimation of vanilla knowledge distillation. }

%Where $\mathcal{L}_{\text {train }}$ and $\mathcal{D}_{\text {train }}$ are the training loss and training dataset, respectively. In this way, training resources can be concentrated on the selected promising sub-networks to enhance their model accuracy.

\section{The Accelerator Search} \label{sec:accelerator search}
% In this section, we propose our coarse-to-fine accelerator search strategy that addresses the challenge of an excessively vast search space. Particularly, we first introduce the accelerator configuration and its corresponding accelerator search space; then the coarse-to-fine search strategy which divides the whole search into a coarse search and a fine search to 
% alleviate the search cost; finally, we integrate the hardware search with the neural architecture search introduced in Sec. \ref{sec:nas} to present the complete NAAS-H algorithm. 
% \subsection{Micro-Architecture and Search Space} \label{sec:hardware search space}
\shh{In this section, 
% we propose our coarse-to-fine accelerator search strategy that addresses the challenge of an excessively vast search space. Particularly, 
we first introduce the micro-architecture and search space of our dedicated accelerator; Then we illustrate the proposed coarse-to-fine accelerator search strategy in Sec. \ref{sec:c2f}, which divides the original vast accelerator search space into several smaller ones to enhance accelerator search efficiency; Finally, Sec. \ref{sec:hw_naas} integrates the accelerator search with the previously introduced neural architecture search to unveil the comprehensive NASH framework.}

\subsection{Micro-Architecture and Search Space} \label{sec:hardware search space}
\begin{figure}[t]
    % %\vspace{-1.5em}
    \centerline{\includegraphics[width=\linewidth]{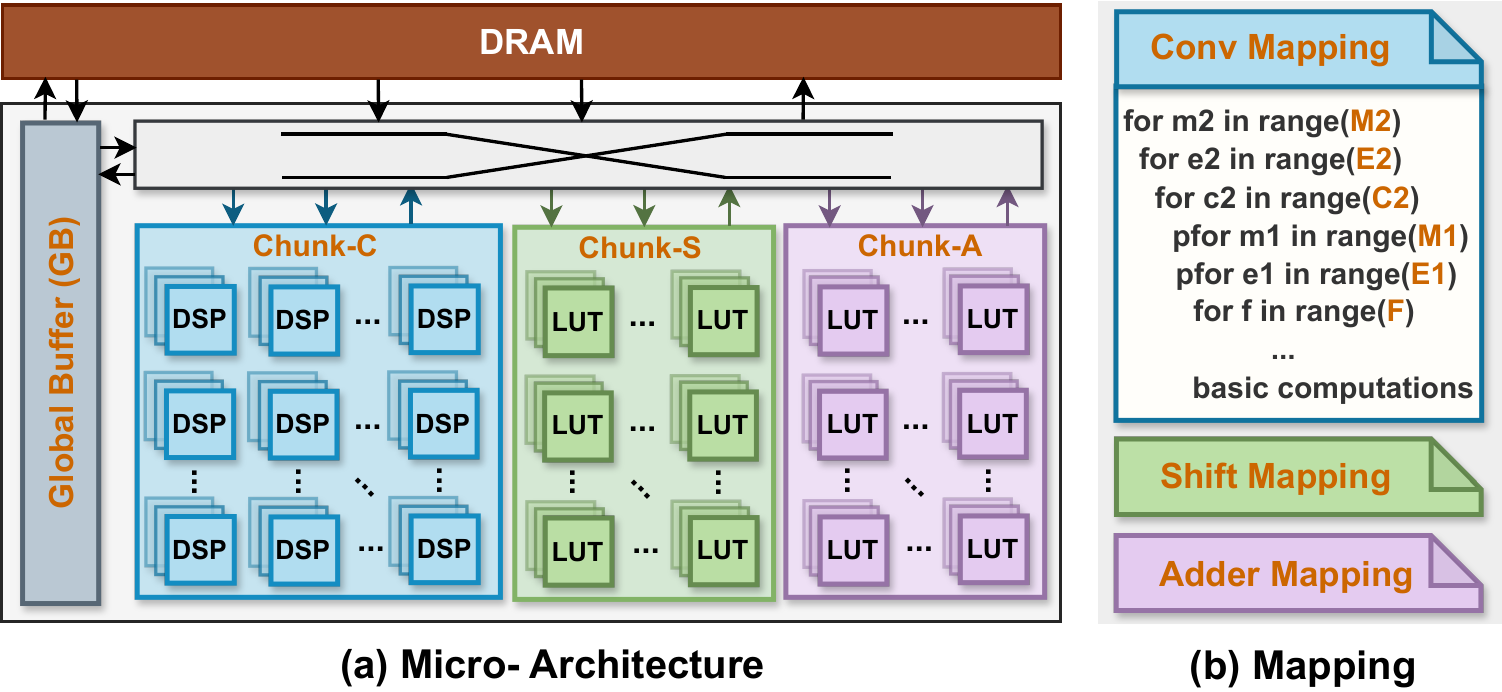}}
	%\vspace{-0.6em}
    \caption{(a) depicts the accelerator micro-architecture, which leverages distinct hardware resources on FPGAs to develop tailored chunks, dubbed Chunk-C, Chunk-S, and Chunk-A, to support convolutions, shift, and adder layers within searched hybrid models, respectively.
    (b) shows mapping methods (i.e., dataflows) for chunks using the widely adopted for-loop description \cite{eyeriss}.
    Notably, components labeled in orange denote the searchable elements within our accelerator search space.
% The accelerator configuration and mapping method. To utilize the distinct hardware resources on FPGAs, our accelerator advocates multiple tailored chunks dubbed Chunk-C, Chunk-S, and Chunk-A to support convolutions, shift, and adder layers in searched hybrid models, respectively. Additionally, the mapping demonstrates how computations of convolution, shift, and adder layers are scheduled on the accelerator. Notably, the orange parts indicate the configurable part that will be optimized in our accelerator search.
}
    %The micro-architecture of our chunk-based accelerator.
    %Specifically, it advocates multiple tailored chunks dubbed Chunk-C, Chunk-S, and Chunk-A equipped with \emph{multiplication} and accumulation (MAC), \emph{shift} and accumulation (SAC), \emph{addition} and accumulation (AAC) units to support convolutions, shift, and adder layers in searched hybrid models, respectively.

	\label{fig:accelerator} %\vspace{-0.7em}
\end{figure}
% %\vspace{-0.7em}

\textbf{Micro-architecture.} 
% To better support the heterogeneous hybrid models, our accelerators roughly align with the design principles of NASA-F \cite{nasa-f}, utilizing a multi-chunk design on FPGA. 
% As illustrated in Fig. \ref{fig:accelerator}, our accelerators mainly comprise an off-chip DRAM, an on-chip Global Buffer (GB), and three distinct chunks, namely Chunk-C, Chunk-S, and Chunk-A. 
% Notably, to enhance hardware utilization and overall throughput, we leverage the diverse resources on FPGA to build customized processing elements (PEs) for each chunk. Specifically, the PEs in Chunk-C are tailored for processing convolution layers and are supported by Digital Signal Processors (DSPs) on FPGA, while the PEs in Chunk-S and Chunk-A are designed for shift layers and adder layers, respectively, and are constructed based on Look-Up Tables (LUTs). 
% Additionally, the PEs in Chunk-C, Chunk-S, and Chunk-A are denoted as MACs (multiplication and accumulations), SACs (shift and accumulations), and AACs (addition and accumulations), respectively.
% Different from NASA-F, to enable accelerator search, the PE number and computation mapping for each chunk, along with the overall GB size, are configurable.
\shh{
To support our multiplication-reduced hybrid models, our dedicated accelerator advocates a chunk-based design, which incorporates several tailored chunks to independently process heterogeneous layers in multiplication-reduced hybrid models following \cite{NASA, nasa-f}. 
As illustrated in Fig. \ref{fig:accelerator} (a), our accelerator mainly comprises an off-chip DRAM, an on-chip Global Buffer (GB), and three distinct chunks (i.e., sub-processors), dubbed Chunk-C, Chunk-S, and Chunk-A. 
Particularly, to enhance hardware utilization and overall throughput, we follow NASA-F \cite{nasa-f} to employ diverse computing resources available on FPGAs to build customized processing elements (PEs) within each chunk. For instance, PEs in Chunk-C are built by Digital Signal Processors (DSPs), aiming to effectively support multiplications in convolutions.
In contrast, PEs in Chunk-S/Chunk-A are developed via Look-Up Tables (LUTs) to efficiently handle bit-wise shifts/additions in shift/adder layers. 
% Additionally, the PEs in Chunk-C, Chunk-S, and Chunk-A are denoted as MACs (multiplication and accumulations), SACs (shift and accumulations), and AACs (addition and accumulations), respectively.
% Different from NASA-F, to enable accelerator search, the PE number and computation mapping for each chunk, along with the overall GB size, are configurable.
}

% \begin{table}[]
% \centering
% \caption{The accelerator search space}
% \resizebox{0.7\linewidth}{!}{
% \setlength{\tabcolsep}{0.35em}
% \begin{tabular}{c|ccc}
% \hline \hline
% Components    & \multicolumn{3}{c}{Searchable Parameters} \\ \hline \hline
% Chunk-C       & PE Number  & Tiling Size  & Tiling Order  \\ 
% Chunk-S       & PE Number  & Tiling Size  & Tiling Order  \\ 
% Chunk-A       & PE Number  & Tiling Size  & Tiling Order  \\ 
% Global Buffer & \multicolumn{3}{c}{Buffer Size}           \\ \hline \hline
% \end{tabular}} \label{tab:hardware search space} %\vspace{-1em}
% \end{table}

\textbf{Search Space.} 
% Formally, our accelerator search space is presented in Table. \ref{tab:hardware search space}. Within this accelerator search space, we encompass both hardware configurations and computation mapping methods. To be more precise, the hardware configuration is defined by parameters such as PE number and GB size, whereas the mapping search space incorporates parameters such as tiling order and tiling size.
% For PE number, buffer size, and tiling size, our search space encompasses all possibilities that FPGA resources allow. As for tiling order, there are four distinct tiling orders derived from popular dataflows, namely, weight stationary (ws), output stationary (os), input stationary (is), and row stationary (rs) dataflows, offering extensive exploration opportunities.
% However, due to the individual searchable parameters for each heterogeneous chunk, the search space within our dedicated accelerators for multiplication-reduced models becomes exponentially intricate. Precisely, compared to the search space of a single-chunk accelerator for multiplication-based models, the search space of our multi-chunk accelerator is about $1.6\times10^5$ large, calling for more effective search strategies.
\shh{To enable accelerator search, the (i) \textit{hardware configuration}, including the PE number of each chunk and buffer size of GB, as well as (ii) \textit{the mapping method} (i.e., dataflow), are searchable in our accelerator. 
%Specifically, as for the {PE number} and {GB size}, we search through all possible choices under the resource budget.
Specifically, 
% as for the {PE number} and {GB size}, all possible choices under the resource budget are encompassed in our search space.
as for the search space of dataflow (see Fig. \ref{fig:accelerator} (b)), we leverage the widely adopted nested for-loop description \cite{eyeriss}, which is characterized by loop ordering factors and loop tiling size. Among them, (i) the former describes the scheduling of computations among the PE array and within each PE, thereby determining data reuse patterns. To enhance search efficiency while maintaining generality, we search from four representative loop orders, including weight stationary (ws), output stationary (os), input stationary (is), and row stationary (rs), for each chunk. Hence, there are a total of $4 \times 4 \times 4 = 64$ combination patterns of mapping methods for our chunk-based accelerator, wherein three dedicated chunks independently handle convolutions, shift layers, and adder layers. 
(ii) Regarding the loop tiling size, it dictates how data are stored within each memory hierarchy to align with the specified loop tiling factors. It can be derived from all feasible choices within the given resource budget and model size constraint.}

% Regarding the potential options for each parameter within the search space, for \textbf{tiling order}, we search from four distinct tiling orders derived from popular dataflows, namely, weight stationary (ws), output stationary (os), input stationary (is), and row stationary (rs) dataflows. %Consequently, we have a total of 64 choices for the reuse pattern of the three chunks in our accelerator. 
% As for the \textbf{tiling size}, the possible choices for a certain loop’s size encompass all the choices that the corresponding data dimension can be factorized into. 
% For the \textbf{PE number} and \textbf{GB size}, we search through all possible choices under the resource budget.

\subsection{Coarse-to-Fine Search Strategy}
\label{sec:c2f}
% As discussed in Sec. \ref{sec:hardware search space}, the existence of multiple heterogeneous chunks exponentially expands the accelerator search space. Since directly managing such an extensive search space is not efficient, we wonder if we can break down this space into several smaller, more manageable ones, and address them sequentially. To answer this question, we draw several observations derived from the direct exploration of the vast search space.
\shh{\textbf{Motivation}. As discussed above, due to the existence of multiple chunks in our dedicated accelerator, the accelerator search space is \textit{exponentially} expanded. Specifically, the combination choices of PE numbers alongside those of mapping methods are exponentially increased with the number of chunks, making it non-trivial to identify the optimal solution from such an enormous search space, thus calling for effective search strategies.}

\begin{figure}[t]
    % %\vspace{-1.5em}
    \centerline{\includegraphics[width=0.9\linewidth]{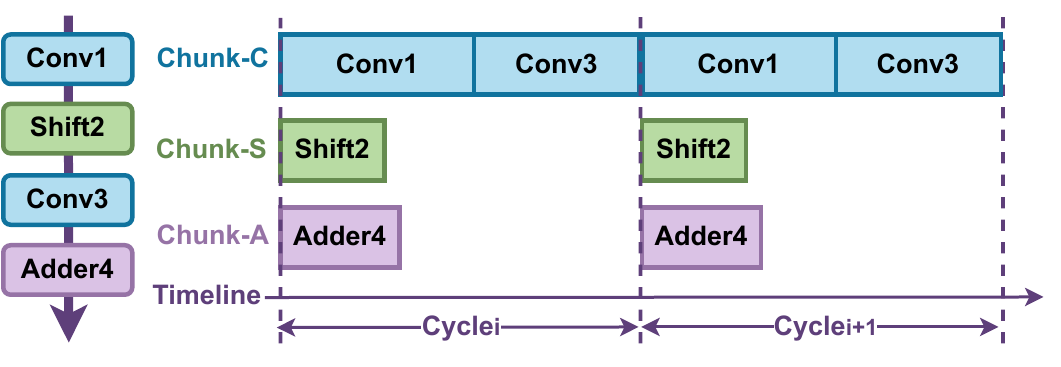}}
	%\vspace{-0.6em}
    \caption{Illustrating the processing timeline of our chunk-based accelerator, where we use a hybrid model consisting of four layers (Conv1, Shift2, Conv3, and Adder4) as an example. 
    % Layers are identified by their layer type and index, such as Conv1 denoting the convolution in the first layer. 
    % The execution time of each layer is an illustrative example after exhaustively leveraging all DSP slices and LUTs to construct PEs. This unbalanced latency among heterogeneous chunks indicates the latency-dominant nature of Chunk-C because of the limited DSP resources.
    Due to the limited DSP resources available on FPGAs, DSP-based Chunk-C emerges as the most latency-dominated chunk in our accelerator.}
    %The micro-architecture of our chunk-based accelerator.
    %Specifically, it advocates multiple tailored chunks dubbed Chunk-C, Chunk-S, and Chunk-A equipped with \emph{multiplication} and accumulation (MAC), \emph{shift} and accumulation (SAC), \emph{addition} and accumulation (AAC) units to support convolutions, shift, and adder layers in searched hybrid models, respectively.

	\label{fig:timeline} %\vspace{-0.7em}
\end{figure}

\textbf{Processing Timeline.}
\shh{
To enhance the comprehension of our upcoming proposed solution, we first introduce the processing timeline of our chunk-based accelerator. Particularly, data consumed by each chunk are independent within each cycle and are derived from different input images to facilitate the concurrent processing of chunks in our accelerator \cite{shen2017maximizing, nasa-f, NASA}. Fig. \ref{fig:timeline} employs a hybrid model comprising four layers as an illustrative example.
As we can see, chunks in our accelerator, i.e., Chunk-C, Chunk-S, and Chunk-A, sequentially process their assigned layers, i.e., convolution layers (Conv1 and Conv3), shift layers (Shift2), and adder layers (Adder4), respectively. 
The outputs generated by each cycle will serve as input for the next cycle. To elaborate, the output of Conv1 managed by Chunk-C in $\text{Cycle}_\text{i}$ will become the input for Shift2, which is then computed by Chunk-S in the subsequent $\text{Cycle}_\text{i+1}$. A cycle concludes once all chunks complete processing. Therefore, the overall latency is dominated by the chunk that consumes the most time.}

\textbf{Observation and Our Proposed Solution.}
\shh{As introduced in Sec. \ref{sec:hardware search space}, our accelerator incorporates three dedicated chunks, which leverage distinct computing resources available on FPGAs to independently support heterogeneous layers in our multiplication-reduced hybrid models, aiming to enhance resource utilization. Specifically, Chunk-C leverages DSP slices to effectively process multiplications in convolutions, while Chunk-S/Chunk-A utilizes LUTs to efficiently handle bit-wise shifts/additions in shift/adder layers. However, DSP slices are generally more resource-constrained compared to LUTs on FPGAs. For instance, on the widely-used embedded FPGA, Kria KV260, the number of LUTs surpasses that of DSP slices by approximately $100\times$. This discrepancy establishes \textit{Chunk-C} as the most resource-constrained component.
Nevertheless, as described in the above paragraph, the overall latency of our chunk-based accelerator is predominantly dominated by the chunk with the longest processing time. Consequently, the limited availability of DSP resources on the FPGA designates \textbf{\textit{Chunk-C}} as the latency-bottleneck chunk.}
\shh{Motivated by this observation, we innovatively introduce a coarse-to-fine accelerator search strategy, aiming to slim the search space and facilitate effective search. Firstly, we focus on identifying the optimal hardware configuration (i.e., PE number) and mapping method for the latency-dominated Chunk-C in a coarse granularity. Subsequently, we refine other searchable parameters in a fine granularity.
By doing this, the original vast search space can be effectively partitioned into several smaller ones, which can be then sequentially explored. Thus, the search space is significantly slimmed and the search process is considerably expedited. Next, we will elaborate on our proposed coarse and fine search phases in detail.}

\textbf{Coarse Search. } 
%This observation is consistent in our experiments and enables us to proactively identify configurations that are most favorable for the Chunk-C through a prior coarse search. 
% As outlined in the above observations, Chunk-C is consistently the latency-dominant chunk. To alleviate the bottleneck brought by Chunk-C and enhance the overall latency, in the coarse search, we search for the optimal PE number, tiling size, and tiling order for Chunk-C under the maximum resource budget, ignoring the computation in other components. For other parameters that are temporarily not included during the coarse search, GB size is set to its maximum value while the parameters for Chunk-S and Chunk-A are not involved during the coarse search. To identify the optimal hardware configuration and mapping method for Chunk-C, we first directly allocate the max available PE number to Chunk-C to guarantee its optimal performance instead of searching through other possible choices. Then for the tiling order and tiling size, we leverage a naive iteration-based search that iterates through all possible choices to identify their optimal values. Although prior works tend to use complex algorithms to avoid this iteration process \cite{DNA,NAAS}, they can not guarantee to find the optimal results. On the other hand, considering the relatively small search space in our coarse search, the naive iteration-based search can also be performed very swiftly while ensuring optimal results, making it a suitable algorithm for our decomposed search space.
\shh{As analyzed above, due to the limited DSP resources on FPGAs, Chunk-C is the most latency-dominated chunk in our accelerator. To alleviate this bottleneck and enhance overall throughput, we first solely identify the optimal PE number and dataflow (including loop ordering factors and loop tiling size) for Chunk-C via the coarse search phase. Specifically, (i) as for the PE number, rather than exhaustively exploring all potential choices, we opt to directly set it to the maximum available value. This decision is driven by the fact that the available DSP resources, and consequently the allowable PE number in Chunk-C, have a direct impact on overall latency. Thus, this tailored handcrafted setting can ensure optimal performance for Chunk-C while mitigating the associated search cost. 
(ii) Regarding the tiling order and tiling size, we systematically iterate through all possible choices to identify the optimal dataflow. It is noteworthy that while previous works \cite{DNA, NAAS} have employed complex algorithms to expedite this iteration process, they fall short of guaranteeing optimal results. Fortunately, due to our proposed coarse-to-fine search strategy and the resultant relatively small search spaces, the straightforward iteration-based search proves not only fast but also ensures optimal performance, aligning well with our coarse search phase. 
(iii) For other parameters excluded from the coarse search, the GB size is configured to its maximum available value to ensure optimal performance for Chunk-C, while the exploration of Chunk-S and Chunk-A is temporarily deferred for subsequent refinement. }

\textbf{Fine Search. } 
% For the subsequent fine search, we involve the computation in Chunk-S and Chunk-A, and search the components that remain unexplored during the coarse search (i.e., Chunk-S, Chunk-A, GB).
% However, although the search space in the fine search is already slimmed, it is more extensive than the search space in the coarse search, challenging the deployment of the iteration-based search method. To further accelerate the search process while ensuring optimal results, we simplify some parameters' search according to the property of our accelerator. Specifically, for the PE number in Chunk-S and Chunk-A, as illustrated in Fig. \ref{fig:timeline}, exhaustively utilizing the LUTs to construct PEs for Chunk-S and Chunk-A does not accelerate the overall processing since Chunk-C dominates the latency, but brings additional hardware resource consumption. Therefore, to roughly balance the execution time of each chunk, similar to NASA and NASA-F \cite{NASA,nasa-f}, we initialize the PE numbers for Chunk-S and Chunk-A based on the PE number for Chunk-C and the respective operation counts of convolution, shift, and adder layers:
\shh{
After selecting the optimal PE number and dataflow for Chunk-C, we proceed to refine other searchable parameters via the fine search phase. This involves searching for PE numbers and mapping methods for both Chunk-S and Chunk-A, along with determining the buffer size of GB.
It can be easily observed that although the search space is significantly slimmed owing to our proposed coarse-to-fine search strategy, the fine search phase still encompasses a more extensive search space than the coarse search, challenging the feasibility of employing the iteration-based search method. 
To overcome this challenge and facilitate an effective search, we further streamline the search space associated with this fine search phase by capitalizing on the inherent hardware characteristics of our accelerator. 
Specifically, (i) as for PE numbers in Chunk-S and Chunk-A, considering Chunk-C dominates latency (see Fig. \ref{fig:timeline}), excessive use of LUTs to construct redundant PEs for both Chunk-S and Chunk-A cannot expedite overall processing but rather incurs additional hardware resource consumption. Guided by this insight, we initialize PE numbers for Chunk-S ${N_\text{Chunk-S}}$ and Chunk-A ${N_\text{Chunk-A}}$ based on the predetermined PE number for Chunk-C ${N_\text{Chunk-C}}$ as well as the operation numbers of convolutions $O_\text{Conv}$, shift layers $O_\text{Shift}$, and adder layers $O_\text{Adder}$:}
\begin{equation}
    \begin{aligned}
        &\frac{N_\text{Chunk-C}}{O_\text{Conv}}=\frac{N_\text{Chunk-S}}{O_\text{Shift}}=\frac{N_\text{Chunk-A}}{O_\text{Adder}},
        \end{aligned} \label{eq:pe-allcation} % %\vspace{-0.3em}
\end{equation}
% where ${N_\text{Chunk-C}}$, ${N_\text{Chunk-S}}$, ${N_\text{Chunk-A}}$ denote the PE number for Chunk-C, Chunk-S, and Chunk-A, respectively and $O_\text{Conv}$, $O_\text{Shift}$, $O_\text{Adder}$ are the operation count numbers of convolution, shift, adder layers. 
% After this initialization, we only need to refine them during the fine search instead of iterating their all possible choices. Additionally, the GB size can be derived by calculating the minimum buffer size needed for the computation schedules of the three chunks. Consequently, we can initially set the GB size to its maximum value during the fine search, and directly obtain its optimal value once the computation schedules are determined. On top of these simplifications, we are able to perform an iteration-based search for Chunk-S and Chunk-A's mapping methods effectively.
\shh{which means that the allocation of PEs to each chunk is determined by the operation number of its assigned layers, aiming to achieve a balanced execution time across chunks \cite{nasa-f, NASA}.
Based upon this initialization, we only need to finetune ${N_\text{Chunk-A}}$ and ${N_\text{Chunk-A}}$ instead of iterating through all possible choices, thus significantly reducing search costs while preserving search accuracy. 
(ii) Thanks to the aforementioned simplification, we then allocate the saved search resources to thoroughly explore mapping methods for Chunk-S and Chunk-A, thus ensuring optimal dataflow.
(iii) Finally, once PE numbers and mapping methods are established for all chunks, the buffer size of the GB is directly calculated as the minimum size required for computations \cite{hong2023dosa} rather than exhaustive searching, increasingly facilitating the search process.
% Consequently, we can initially set the GB size to its maximum value during the fine search, and directly obtain its optimal value once the computation schedules are determined.
}

\begin{figure}[t]
    % %\vspace{-1.5em}
    \centerline{\includegraphics[width=\linewidth]{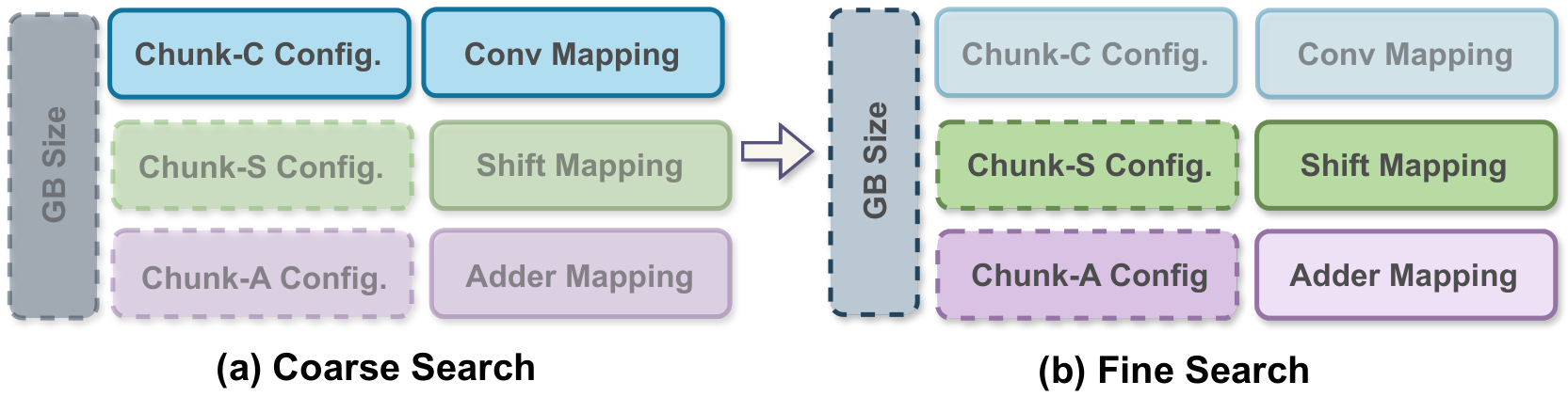}}
	%\vspace{-0.6em}
    \caption{
    % Illustration of our coarse-to-fine search strategy. In (a) Coarse search, we identify the most performance-dominant component in our search space (i.e., Chunk-C) and concentrate our hardware resources to search for its optimal configuration. In (b) Fine search, we explore the rest components, achieving the optimal result and minimizing resource consumption. Notably, the PE number of Chunk-S and Chunk-A is computed in Coarse search and fine-tuned in Fine search.
    Illustrating our coarse-to-fine accelerator search strategy. (a) In the coarse search, we first optimize the bottleneck Chunk-C. Subsequently, (b) in the fine search, we refine other searchable components, including Chunk-S, Chunk-A, and global buffer (GB).
    }
	\label{fig:c2f} %\vspace{-0.7em}
\end{figure}
%\vspace{-0.7em}

\subsection{Neural Architecture and Accelerator Co-Search}
\label{sec:hw_naas}
% In this section, we integrate the accelerator search into the neural architecture search introduced in Sec. \ref{sec:neural architecture search}. Specifically, when evaluating a network's performance, besides assessing its algorithmic performance by Eq. (\ref{eq:score}), we also perform an accelerator search for this network and obtain its hardware feedback. Formally, we present our NAAS-H algorithm in Alg. \ref{alg:NASA-H}, which incorporates both neural architecture search and accelerator search to fully explore the joint optimization space.
\shh{By integrating the aforementioned accelerator search methodology into the previously introduced neural architecture search, we successfully derive our NASH neural architecture and accelerator search framework. Specifically, in the process of identifying promising multiplication-reduced sub-networks, we evaluate the algorithmic performance of models using our tailored zero-shot metric, as defined in Eq. (\ref{eq:score}). Furthermore, we estimate the optimal hardware performance of these models employing our proposed coarse-to-fine accelerator search strategy. Formally, we outline the computation pipeline of our NASH framework in Alg. \ref{alg:NAAS-H}, which seamlessly integrates both neural architecture search and accelerator search, aiming to directly identify optimal pairing of multiplication-reduced hybrid models and dedicated accelerators.
% exhibiting both promising algorithmic potential and optimal hardware performance.
}

{
\begin{figure}[!t]

\makeatletter
\newcommand{\removelatexerror}{\let\@latex@error\@gobble}
\makeatother

\begin{minipage}{0.477\textwidth}
\centering

\begingroup
\removelatexerror% Nullify \@latex@error

\begin{algorithm}[H]

% \caption{ViTCoD Split and Conquer Algorithm.}
\caption{NASH: Neural Architecture and Accelerator Co-Search for Multiplication-Reduced Models}
\label{alg:NAAS-H}

\SetAlgoLined
\KwIn{Hybrid model search space, accelerator search space, and hardware constraints} 
\KwOut{Optimal network and dedicated accelerator pairing}
% $A[A<\theta_p]\leftarrow0$\label{algl:prune}\Comment*[r]{prune small values in $A$}

% pruning

\CommentL{\textcolor{dark-purple}{{\textit{// Evolutionary Search}}}}
Randomly sample a population of sub-networks $\mathcal{A}$

\For{$iter=1$ \KwTo $max \ iteration$}{
    Expand the population $\mathcal{A}$ by crossover and mutation
    
    \For{each network $\alpha$ in $\mathcal{A}$}{
    \CommentL{\textcolor{dark-blue}{{\textit{// Hardware Evaluation}}}}
    
    \tikzmka{A}
    Employ coarse-to-fine search following Sec. \ref{sec:c2f} to obtain the \textbf{\textit{best accelerator}} for network $\alpha$

    Measure hardware performance for network $\alpha$ \ \ \ \ \ \ \ \ \ \ \ \
    \tikzmkc{B} \boxita{light-blue!65}
    
    \CommentL{\textcolor{dark-orange}{{\textit{// Algorithm Evaluation}}}}
    
    \tikzmkc{A}
    Compute $n  =  \texttt{NN{-}Degree}(\alpha)$ via Eq. (\ref{eq:nn-degree})

     Compute $z =  \texttt{Zen-Score}(\alpha)$ via Eq. (\ref{eq:zenscore}) \ \ \ \ \ \ \ \ \ \ \ \ \ \ \ \ \ \
    \tikzmkc{B} \boxitc{orange}

    }
    
    \tikzmkc{A}
     Compute zero-shot score $s$ using $z$ and $n$ via Eq. (\ref{eq:score})  \ \ \ \ \ \ 
    \tikzmkc{B} \boxitc{orange}
    
     Update $\mathcal{A}$ by selecting \textbf{\textit{networks with top-k $\mathbf{s}$}} under given hardware constraints
}

Do preference-biased supernet training via Eq. (\ref{eq:training})

\end{algorithm}

\endgroup

\end{minipage}
%\vspace{-1.5em}
\end{figure}
}

\section{Experiments} \label{sec:experiments}
% In this section, we first clarify our experimental setup in Sec. \ref{sec:algo-set-up}, then compare our NAAS-H framework with SOTA systems in Sec. \ref{sec:oversota}. Next, we validate the effectiveness of our zero-shot neural architecture search and accelerator search in Sec. \ref{sec:exp_alg} and Sec. \ref{sec:exp_hw}, respectively.
\shh{In this section, we first clarify our experimental setup, then compare our NASH framework with SOTA systems in Sec. \ref{sec:oversota}. Finally, we validate the effectiveness of our zero-shot architecture search and coarse-to-fine accelerator search enablers in Sec. \ref{sec:exp_alg} and Sec. \ref{sec:exp_hw}, respectively.}

%\vspace{-0.8em}
\subsection{Experimental Setup}
\label{sec:algo-set-up}

\paragraph{\textbf{Datasets, Baselines, and Evaluation Metrics}}
% To validate our NAAS-H framework which incorporates 
% %(i) a neural architecture and accelerator co-search process empowered by a tailored zero-shot metric and a coarse-to-fine accelerator search
% (i) the zero-shot neural architecture search to pre-identify promising sub-networks and thus improve model accuracy as introduced in Sec. \ref{sec:neural architecture search} and (ii) the coarse-to-fine accelerator search to boost the throughput and hardware utilization, 
% we conduct experiments on three datasets, including CIFAR10, CIFAR100, and Tiny-Imagenet.
\shh{To validate our NASH framework, which incorporates (i) a \textit{zero-shot architecture search} to pre-identify promising multiplication-reduced models, thus alleviating gradient conflict during the subsequent preference-biased supernet training with boosted accuracy, and (ii) a \textit{coarse-to-fine accelerator search} to partition and slim the enormous accelerator search space for accelerating the accelerator search process, we conduct experiments on \underline{\textbf{\textit{three datasets}}}, including CIFAR10, CIFAR100, and Tiny-Imagenet.}
% We compare our NAAS-H with \underline{\textbf{\textit{six baselines}}}, including \textbf{\emph{multiplication-based systems}}: (i) the SOTA multiplication-based one-shot NAS work AlphaNet \cite{AlphaNet}, (ii) the SOTA multiplication-based zero-shot NAS work PreNAS \cite{wang2023prenas};
% SOTA \textbf{\emph{multiplication-free systems}}: (iii) handcrafted AdderNet \cite{AdderNet} and (iv) DeepShift \cite{DeepShift};
% as well as \textbf{\emph{multiplication-reduced system}}: (v) the SOTA multiplication-reduced one-shot supernet-based NAS work NASA-F \cite{nasa-f}. For the deployment, we build dedicated FPGA-based accelerators for each baseline following the design principles outlined in NASA-F \cite{nasa-f}. Specifically, these baseline accelerators' mapping methods and hardware configurations are fixed or pre-determined since no accelerator search is performed.
\shh{We compare our NASH with \underline{\textbf{\textit{five baselines}}}, including SOTA \textbf{\emph{multiplication-based systems}}: multiplication-based models searched by the SOTA multiplication-based (i) one-shot NAS work AlphaNet \cite{AlphaNet} and (ii) zero-shot search work PreNAS \cite{wang2023prenas} and executed on the dedicated FPGA-based accelerator with SOTA DSP-implementations \cite{Xilinx-conv} following \cite{nasa-f};
SOTA \textbf{\emph{multiplication-free systems}}, such as (iii) handcrafted multiplication-free models AdderNet \cite{AdderNet} and (iv) DeepShift \cite{DeepShift} executed on their dedicated FPGA accelerators with hardware-efficient LUT-implementations \cite{nasa-f};
as well as the SOTA \textbf{\emph{multiplication-reduced system}}, i.e., (v) multiplication-reduced models searched by the SOTA multiplication-reduced one-shot supernet-based NAS work NASA-F \cite{nasa-f} and accelerated on its dedicated chunk-based accelerator. 
% For the deployment, we build dedicated FPGA-based accelerators for each baseline following the design principles outlined in NASA-F \cite{nasa-f}. Specifically, these baseline accelerators' mapping methods and hardware configurations are fixed or pre-determined since no accelerator search is performed.
}
% \textcolor{red}{For the hardware performance of other baselines, we obtained the results of AlphaNet \cite{AlphaNet}, NASA-F \cite{nasa-f}, and multiplication-free baselines directly from previous work \cite{nasa-f}. Regarding PreNAS \cite{wang2023prenas}, which focuses solely on algorithmic performance and does not include hardware results in its paper, we follow NASA-F \cite{nasa-f} to assess the hardware performance of its searched models on a developed baseline accelerator, which exhibits a similar hardware architecture to ours. The primary difference is that our dedicated accelerator is equipped with three chunks to independently process convolutions, shift layers, and adder layers within our searched hybrid multiplication-reduced network. In contrast, the baseline accelerator has only one chunk for processing convolutions solely involved in the multiplication-based models searched by PreNAS. Additionally, to ensure fair comparisons, we maintain consistency in clock frequency, bit-width, DSP number, and DSP implementation strategy across both our and baseline accelerators.}
% Moreover, we evaluate the performance of our NAAS-H and other baselines in terms of the following four metrics: accuracy (top-1 accuracy by default), frame rate (i.e.,  Frame-Per-Second, FPS), overall throughput, and hardware utilization efficiency. 
\shh{Moreover, we compare our NASH with other baselines in terms of \underline{\textbf{\textit{four evaluation metrics}}}: accuracy (top-1 accuracy by default), frame rate (Frame-Per-Second, FPS), overall throughput, and hardware utilization efficiency.}

\paragraph{\textbf{Search Setup}}
% Our NAAS-H integrates both neural architecture search and accelerator search. 
% For the zero-shot \textbf{neural architecture search}, we employ an evolutionary algorithm, where the population size is set to be $100$, the size and probability of mutation/crossover are set to be $50$ and $0.2$, respectively, and we run the evolutionary search for $15$ iterations. 
% %where the score derived from the tailored zero-shot metric and the hardware feedback obtained from the searched accelerator serve as the metrics for evolution.
% %which integrates both neural architecture search and accelerator search. 
% % For the \textbf{evolutionary algorithm settings}, 
% %Since both the tailored zero-shot metric and the coarse-to-fine accelerator search are time-efficient, 
% In each iteration, the top-N scoring sub-networks in each group are kept as introduced in Sec. \ref{sec:The Network Search Strategy}, and in our experiments, N is set to be $3$.
\shh{Our NASH integrates both neural architecture search and accelerator search. 
For the zero-shot \underline{\textit{\textbf{architecture search}}}, as introduced in Sec. \ref{sec:nas}, to expedite the search process, we employ an evolutionary algorithm, where the population size is set to be $100$, and the size and probability of mutation/crossover are set to be $50$ and $0.2$, respectively. We run the evolutionary search for $15$ iterations, and only top-3 sub-architectures are retrained during each iteration (i.e., $k=3$ in our experiments). }

% For the \textbf{accelerator search}, we construct our accelerators on Xilinx Kria KV260 embedded FPGA platform, which consists of $1248$ DSP slices and $117$k LUT resources under a clock frequency of $200$MHz. 
% %All hardware designs are implemented on Vivado 2022.2 HLS design flow \cite{Xilinx} under the clock frequency of $200$MHz. 
% During the accelerator search, we follow NASA \cite{NASA} and NASA-F \cite{nasa-f} to build a cycle-accurate chip predictor/simulator to provide fast and reliable estimations, which are validated against the RTL implementation to ensure correctness. 
% Additionally, we also employ the SOTA quantization method \cite{ScalableMF} to quantize both weights and activations into $8$ bits, except weights in shift layers that are quantized into $4$ bit.
\shh{For the \underline{\textit{\textbf{accelerator search}}}, we implement our accelerator on the wildely-adopted Xilinx Kria KV260 embedded FPGA platform, which consists of $1248$ DSP slices and $117$k LUT resources, under a clock frequency of $200$MHz. 
To provide fast and reliable estimations, we follow \cite{NASA, nasa-f} to build a cycle-accurate chip predictor/simulator upon our accelerator and validate it against the RTL implementation to ensure correctness. 
Additionally, we also follow \cite{NASA, nasa-f} to quantize both weights and activations into $8$ bits, except weights in shift layers that are quantized into $4$ bit.}

\paragraph{\textbf{Training Recipes}}
% For the one-shot supernet-based training, we consistently optimize the sub-networks derived from the search process by the following recipes: (i) for CIFAR-10 and CIFAR-100, we optimize these sub-networks for $800$ epochs, with the help of sandwich rule-guided sampling method \cite{BigNAS} and $\alpha$-divergence-based knowledge distillation \cite{AlphaNet} to facilitate the supernet training. We also conduct our warm-up strategy in the first $30$ epochs of training. For the optimization of weights, we use the SGD optimizer with a momentum of $0.9$, and the initial learning rate (lr) is $0.3$, which decays following a cosine scheduler while the batch size is set to $384$. (ii) For the more challenging Tiny-ImageNet, the training epoch is reduced to $250$ epochs. Besides, the batch size and learning rate are adjusted to $128$ and $0.1$, respectively.
%mark
\shh{%After identifying promising multiplication-reduced sub-architectures via our proposed zero-shot architecture search and coarse-to-fine accelerator search, we can then allocate training resources to optimize these derived networks.
As introduced in Sec. \ref{sec:training}, optimization of identified sub-architectures is accomplished using the SOTA one-shot supernet-based training strategy \cite{nasa-f, AlphaNet}, which incorporates the sandwich rule-guided sampling method \cite{BigNAS} and $\alpha$-divergence-based knowledge distillation \cite{AlphaNet} to facilitate the supernet training.} 
% Specifically, (i) for CIFAR-10 and CIFAR-100, we optimize these sub-networks for $800$ epochs, using the SGD optimizer with a momentum of $0.9$, and the initial learning rate (lr) is $0.3$, which decays following a cosine scheduler while the batch size is set to $384$. (ii) For the more challenging Tiny-ImageNet, the training epoch is reduced to $250$ epochs. Besides, the batch size and learning rate are adjusted to $128$ and $0.1$, respectively. The other settings remain the same as in CIFAR10/CIFAR100, i.e., the optimizer is still the SGD optimizer with a momentum of $0.9$ while
% the learning rate decays following a cosine scheduler.
\shh{Specifically, (i) for CIFAR-10 and CIFAR-100, we optimize these sub-networks for $800$ epochs, using the SGD optimizer with a momentum of $0.9$. The initial learning rate (lr) is $0.3$, which decays following a cosine scheduler. The batch size is set to $384$. (ii) For the more challenging dataset Tiny-ImageNet, the training epoch is set to $250$ epochs. Besides, the batch size and learning rate are adjusted to $128$ and $0.1$, respectively. The remaining settings are consistent with those employed in  CIFAR10/CIFAR100.
%mark
%i.e., utilizing the SGD optimizer with a momentum of $0.9$, and using a cosine scheduler to decay the learning rate.
}

\subsection{NASH over SOTA Systems}
\label{sec:oversota}

\begin{figure}[t]
    % %\vspace{-1.5em}
    \centerline{\includegraphics[width=\linewidth]{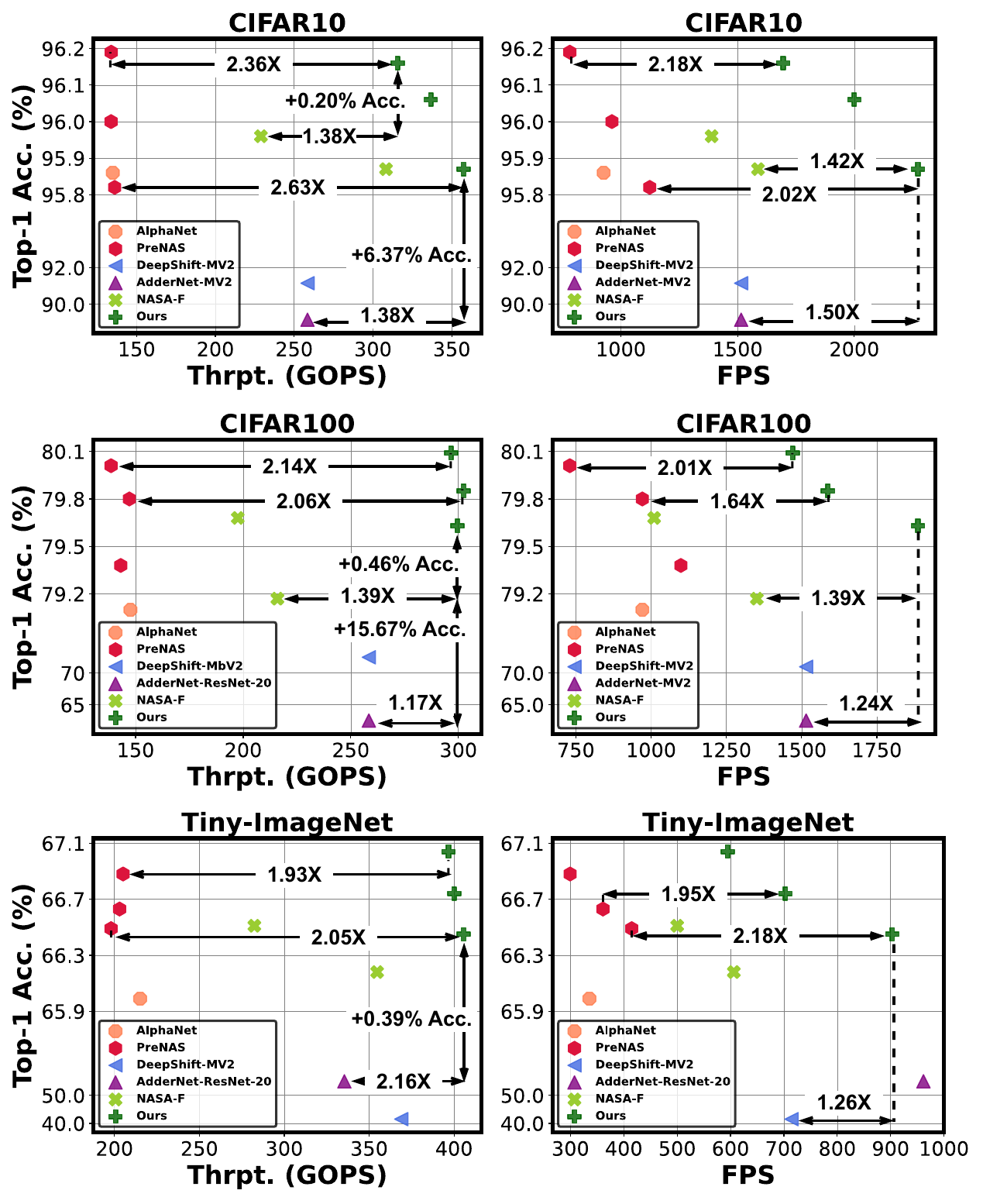}}
	%\vspace{-0.6em}
    \caption{Comparisons regarding throughput (\texttt{Thrpt.}) and {FPS} against accuracy when evaluated on CIFAR10, CIFAR100, and Tiny-ImageNet, respectively. \texttt{MV2} donates the abbreviation of {MobileNetV2} \cite{mobilenetv2} by default.}
	\label{fig:overall_results} %\vspace{-1.2em}
\end{figure}

% As shown in Fig. \ref{}, our NAAS-H framework shows an enhanced trade-off between accuracy and hardware efficiency compared to all SOTA baselines. Specifically, (i) compared to the multiplication-based baselines, our NAAS-H framework significantly outperforms them in hardware efficiency, thanks to the incorporation of multiplication-free layers and the searched multi-chunk accelerator. Meanwhile, NAAS-H also retains comparable and even better accuracy. For instance, we can achieve $?$, $?$, and $?$ throughput improvement as well
% as $?$, $?$, and $?$ better FPS while the accuracy improvement is $?$, $?$, and $?$, compared to the most competitive SOTA zero-shot multiplication-based work PreNAS on the dedicated FPGA-based accelerator with SOTA DSP packing strategy \cite{Xilinx-conv} following \cite{nasa-f},  when tested on CIFAR10, CIFAR100, and Tiny-ImageNet, respectively.
\paragraph{\textbf{Results and Analysis}}
\shh{As shown in Fig. \ref{fig:overall_results}, our NASH framework shows better trade-offs between accuracy and hardware efficiency over all SOTA baselines, demonstrating our effectiveness. Specifically, (i) by incorporating both hardware-efficient multiplication-free layers and powerful convolutions within our searched hybrid models, we significantly outperform SOTA \textbf{\textit{multiplication-based baselines}} in hardware efficiency while maintaining comparable and even better accuracy. Particularly, when compared with the most competitive baseline, 
%mark
% i.e., multiplication-based models searched by the SOTA zero-shot work PreNAS \cite{wang2023prenas} and executed on the dedicated accelerator \cite{nasa-f}, 
i.e., PreNAS \cite{wang2023prenas} on the dedicated accelerator \cite{nasa-f}, 
we can offer \textbf{2.36}$\times$$\sim$\textbf{2.63}$\times$, \textbf{2.06}$\times$$\sim$\textbf{2.14}$\times$, and \textbf{1.93}$\times$$\sim$\textbf{2.05}$\times$ throughput improvements alongside $\mathbf{2.02}\times$$\sim$$\mathbf{2.18}\times$, $\mathbf{1.64}\times$$\sim$$\mathbf{2.01}\times$, and $\mathbf{1.95}\times$$\sim$$\mathbf{2.18}\times$ better FPS with $-0.07\%$$\sim$$0.06\%$, $0.05\%$ $\sim$$0.25\%$, and $-0.04\%$$\sim$$0.16\%$ higher accuracy, when tested on CIFAR10, CIFAR100, and Tiny-ImageNet, respectively.}

% (ii) For the comparison with the multiplication-free systems, NAAS-H presents much higher accuracy due to the existence of convolutions while the hardware efficiency is also enhanced. To be more specifical, we can offer up to $?$, $?$ and 
% $?$ throughput as well as $?$, $?$ and $?$ FPS with $?$, $?$ and $?$ accuracy, compared to handcrafted DeepShift-MobileNet-V2 \cite{DeepShift} on its dedicated accelerator with the efficient LUT implementation \cite{nasa-f} when tested on CIFAR10, CIFAR100 and Tiny-ImageNet, respectively. 
% Moreover, we can also achieve $?$ and $?$ throughput, $?$ and $?$ FPS with $?$ and $?$ accuracy improvement over AdderNet-MobileNet-V2 \cite{AdderNet} on its dedicated accelerator with efficient LUT implementation \cite{nasa-f} when evaluated on CIFAR-10 and CIFAR100. This enhanced hardware efficiency indicates the effectiveness of our searched multi-chunk accelerator.
\shh{(ii) Owing to the existence of convolutions within our searched hybrid models to enhance accuracy as well as the dedicated accelerator incorporated in our NASH framework to boost hardware efficiency, when compared with SOTA \textbf{\textit{multiplication-free systems}}, we can offer much higher accuracy with comparable hardware efficiency. Specifically, we can offer up to $\uparrow$$\mathbf{4.2}\%$, $\uparrow$$\mathbf{9.1}\%$, and $\uparrow$$\mathbf{23.3}\%$ accuracy with $\uparrow$$1.38\times$, $\uparrow$$1.17\times$, and $\uparrow$$1.10\times$ throughput as well as $\uparrow$$1.50\times$, $\uparrow$$1.24\times$, and $\downarrow$$0.93\times$ FPS compared to the handcrafted DeepShift-MobileNet-V2 \cite{DeepShift} executed on its dedicated accelerator \cite{nasa-f} and tested on CIFAR10, CIFAR100, and Tiny-ImageNet, respectively. 
Additionally, we can achieve $\uparrow$$\mathbf{6.7}\%$ and $\uparrow$$\mathbf{16.6}\%$ accuracy with $\uparrow$$1.38\times$ and $\uparrow$$1.17\times$ throughput alongside $\uparrow$$1.50\times$ and $\uparrow$$1.24\times$ FPS, when compared with AdderNet-MobileNet-V2 \cite{AdderNet} on its dedicated accelerator \cite{nasa-f} and evaluated on CIFAR-10 and CIFAR100, respectively.}

% (iii) As for the comparison with SOTA one-shot multiplication-reduced framework NASA-F \cite{nasa-f} on its dedicated multi-chunk FPGA accelerator, NAAS-H also outperforms in terms of both accuracy and hardware efficiency. Precisely, we achieve $?$, $?$, and $?$ throughput improvement as well as $?$, $?$, and $?$ better FPS with the accuracy is $?$, $?$, and $?$ over the most competitive networks in NASA-F, when tested on CIFAR10, CIFAR100, and Tiny-ImageNet, respectively. It is worth noting that the superiority in accuracy validates the effectiveness of the zero-shot NAS concentrating training resources, while the advantage in hardware efficiency is from exhaustively exploring the hardware possibilities through our coarse-to-fine accelerator search.
\shh{(iii) As for comparisons with the SOTA \textbf{\textit{multiplication-reduced baseline}} 
%mark
%, i.e., multiplication-reduced models searched by the SOTA one-shot supernet-based framework NASA-F \cite{nasa-f} and then executed on their dedicated chunk-based accelerator,
NASA-F \cite{nasa-f}, 
NASH also exhibits better accuracy and boosted hardware efficiency. This improved performance underscores the effectiveness of our tailored zero-shot metric in pre-identifying promising multiplication-reduced models, thus facilitating the subsequent supernet training. Additionally, it highlights the superiority of our coarse-to-fine accelerator search in the pursuit of optimal hardware performance.
For instance, we can gain $\uparrow$$1.16\times$$\sim$$\uparrow$$\mathbf{1.38}\times$, $\uparrow$$\mathbf{1.40}\times$$\sim$$\uparrow$$\mathbf{1.50}\times$, and $\uparrow$$1.14\times$$\sim$$\uparrow$$\mathbf{1.40}\times$ throughput alongside $\uparrow$$1.23\times$$\sim$$\uparrow$$\mathbf{1.42}\times$, $\uparrow$$\mathbf{1.39}\times$$\sim$$\uparrow$$\mathbf{1.45}\times$, and $\uparrow$$1.19\times$$\sim$$\uparrow$$\mathbf{1.49}\times$ FPS with $0.19\%$$\sim$ $0.20\%$, $0.41\%$$\sim$$0.46\%$, and $0.53\%$$\sim$$0.56\%$ higher accuracy on CIFAR10, CIFAR100, and Tiny-ImageNet, respectively.}

% \usepackage{multirow}
\begin{table*}[]
\centering
\setlength{\tabcolsep}{3.5pt}
\caption{Comparisons of time costs regarding both supernet-training and search on CIFAR100. 'Thrpt.', 'Acc.', and 'Avg.' are the abbreviations of throughput, accuracy, and average, respectively} 
\renewcommand{\arraystretch}{1.1}
\resizebox{0.95\linewidth}{!}{
\begin{tabular}{c|cc|ccc|c|c}
\Xhline{3\arrayrulewidth}
\multirow{2}{*}{\textbf{Method}} & \multicolumn{2}{c|}{\textbf{Supernet-Training}} & \multicolumn{3}{c|}{\textbf{Search}}                  & \multirow{2}{*}{\textbf{Avg. Thrpt. (GOPS)}} & \multirow{2}{*}{\textbf{Avg. Acc. ($\%$)}} \\ \cline{2-6}
                        & \textbf{Methodology}                  & \textbf{Time Cost} & \textbf{Network}  & \textbf{Hardware}  & \textbf{Total Time Cost} &                                &                                   \\ \hline \hline
AlphaNet \cite{AlphaNet}     & Random Optimization        & 48h       & 6h            & \color{purple}{\xmark}              & 6h        & 147.6                    & 79.1                             \\ \hline
PreNAS \cite{wang2023prenas}     & Concentrated Optimization        & 48h       & 0.5h            & \color{purple}{\xmark}              & 0.5h        & 142.9                    & 79.73                             \\ \hline
NASA-F \cite{nasa-f}                 & Random Optimization        & 84h       & 7h            & \color{purple}{\xmark}              & 7h        & 206.7                    & 79.42                             \\ \hline
\rowcolor{dark-green!15} \textbf{Ours}              & Concentrated Optimization  & 84h       & \textbf{0.5h}            & 5.5h             & 6h        & \textbf{299.5}                   & \textbf{79.86}                             \\ \Xhline{3\arrayrulewidth}
\end{tabular}} \label{table: time and efficiency comparison} 
\end{table*}
% \usepackage{multirow}
\begin{table*}[]
\centering
\setlength{\tabcolsep}{3.5pt}
\caption{Comparisons between ours and SOTA multiplication-based/-free/-reduced baselines regarding operation numbers, {computational energy costs (\texttt{Comp.})}, and accuracy (\texttt{Acc.}), when validated on CIFAR10 and CIFAR100, respectively} 
\renewcommand{\arraystretch}{1.1}
\resizebox{\linewidth}{!}{
\begin{tabular}{c|c|ccccc|ccccc}
\Xhline{3\arrayrulewidth}
\multirow{2}{*}{\textbf{Models}}                                           & \multirow{2}{*}{\textbf{Methods}}  & \multicolumn{5}{c|}{\textbf{CIFAR10}}                                                                 & \multicolumn{5}{c}{\textbf{CIFAR100}}                                                                \\ \cline{3-12} 
                                                                           &                                                                                                                             & \textbf{Multi. (M)} & \textbf{Shift (M)} & \textbf{Add (M)} & \textbf{Comp. (mJ)} & \textbf{Acc. (\%)} & \textbf{Multi. (M)} & \textbf{Shift (M)} & \textbf{Add (M)} & \textbf{Comp. (mJ)} & \textbf{Acc. (\%)} \\ \hline \hline
\multirow{4}{*}{\begin{tabular}[c]{@{}c@{}}Multi.-\\ Based\end{tabular}}   & AlphaNet \cite{AlphaNet}                                                                                                                      & 72.86              & 0.00               & 72.86            & 0.437               & 95.86             & 76.02              & 0.00               & 76.02            & 0.456               & 79.10             \\
                                                                           & PreNAS-A \cite{wang2023prenas}                                                                                                               & 85.46              & 0.00               & 85.46            & 0.513               & 96.19             & 94.82              & 0.00               & 94.82            & 0.569               & 80.01             \\
                                                                           & PreNAS-B \cite{wang2023prenas}                                                                                                              & 69.85              & 0.00               & 69.85            & 0.419               & 96.00             & 75.90              & 0.00               & 75.90            & 0.455               & 79.80             \\
                                                                           & PreNAS-C \cite{wang2023prenas}                                                                                                              & 60.37              & 0.00               & 60.37            & 0.362               & 95.82             & 64.92              & 0.00               & 64.92            & 0.390               & 79.38             \\ \hline
\multirow{2}{*}{\begin{tabular}[c]{@{}c@{}}Multi.-\\ Free\end{tabular}}    & DeepShift-MV2 \cite{DeepShift}                                                                                                           & 6.60               & 79.20              & 85.80            & 0.154               & 91.90             & 6.70               & 79.20              & 85.80            & 0.154               & 71.00             \\
                                                                           & AdderNet-MV2 \cite{AdderNet}                                                                                                           & 6.60               & 0.00               & 165.00           & 0.154               & 89.50             & 6.70               & 0.00               & 165.00           & 0.154               & 63.50             \\ \hline
 & NASA-F-A \cite{nasa-f}                                                                                                                  & 50.87              & 28.53              & 85.42            & 0.351               & 95.96             & 72.48              & 15.68              & 107.34           & 0.471               & 79.68             \\
Multi.- & NASA-F-B \cite{nasa-f}                                                                                                                    & 46.78              & 43.88              & 103.60           & 0.353               & 95.87              & 53.26              & 18.82              & 87.70            & 0.358               & 79.17             \\
 \rowcolor{dark-green!15}        Reduced                                                                      & \textbf{Ours-A}                                                                                                           & \textbf{55.48}     & \textbf{33.81}     & \textbf{95.76}   & \textbf{0.386}      & \textbf{96.16}    & \textbf{71.04}     & \textbf{19.04}     & \textbf{112.02}  & \textbf{0.469}      & \textbf{80.09}   \\
                                             \rowcolor{dark-green!15}                              &  \textbf{Ours-B}                                                                                                           & \textbf{47.97}     & \textbf{33.81}     & \textbf{88.24}   & \textbf{0.341}      & \textbf{96.06}    & \textbf{65.97}     & \textbf{18.64}     & \textbf{105.78}  & \textbf{0.438}      & \textbf{79.85}   \\
                                             \rowcolor{dark-green!15}                              &  \textbf{Ours-C}                                                                                                          & \textbf{42.26}     & \textbf{33.13}     & \textbf{81.85}   & \textbf{0.306}      & \textbf{95.87}    & \textbf{56.61}     & \textbf{14.14}     & \textbf{88.52}   & \textbf{0.373}      & \textbf{79.63}   \\ \Xhline{3\arrayrulewidth}
\end{tabular}} 
\label{tab:alg-exp-cifar} 
\end{table*}
% \usepackage{multirow}
\begin{table}[]
\centering
\setlength{\tabcolsep}{1.5pt}
\caption{Comparisons of operation numbers, {computational energy costs (\texttt{Comp.})}, and accuracy (\texttt{Acc.}) on Tiny-ImageNet} 
\renewcommand{\arraystretch}{1.1}
\resizebox{\linewidth}{!}{
\begin{tabular}{c|c|ccccc}
\Xhline{3\arrayrulewidth}
\multicolumn{1}{c|}{\multirow{2}{*}{\textbf{Models}}}                               & \multicolumn{1}{c|}{\multirow{2}{*}{\textbf{Methods}}} & \multicolumn{5}{c}{\textbf{Tiny-ImageNet}}                                                                                                                           \\ \cline{3-7} 
\multicolumn{1}{c|}{}                                                      & \multicolumn{1}{c|}{}                         & \multicolumn{1}{c}{\textbf{Multi. (M)}} & \multicolumn{1}{c}{\textbf{Shift (M)}} & \multicolumn{1}{c}{\textbf{Add (M)}} & \multicolumn{1}{c}{\textbf{Comp. (mJ)}} & \multicolumn{1}{c}{\textbf{Acc. (\%)}} \\ \hline \hline
\multirow{4}{*}{\begin{tabular}[c]{@{}c@{}}Multi.-\\Based\end{tabular}}   & AlphaNet \cite{AlphaNet}                                                  & 320.43             & 0.00               & 320.43           & 1.923               & 65.99             \\
                                                                         & PreNAS-A \cite{wang2023prenas}                                                   & 342.44             & 0.00               & 342.44           & 2.055               & 66.88             \\
                                                                         & PreNAS-B \cite{wang2023prenas}                                                   & 281.17             & 0.00               & 281.17           & 1.687               & 66.63             \\
                                          & PreNAS-C \cite{wang2023prenas}                                                   & 238.64             & 0.00               & 238.64           & 1.432               & 66.49             \\ \hline
                                                                         & DeepShift-MV2 \cite{DeepShift}                                              & 19.77              & 215.03             & 234.80           & 0.428               & 43.70              \\
\multirow{-2}{*}{\begin{tabular}[c]{@{}c@{}}Multi.-\\ Free\end{tabular}} & AdderNet-Res20 \cite{AdderNet}                                             & 21.23              & 0.00               & 362.45     & 0.373               & 55.30                       \\ \hline

& NASA-F-A \cite{nasa-f}                                                   & 212.96             & 52.28              & 299.61           & 1.378               & 66.51             \\
                                                                      & NASA-F-B \cite{nasa-f}                                                   & 167.59             & 98.40              & 319.25           & 1.186               & 66.18             \\
						\rowcolor{dark-green!15}	Multi.-		& \textbf{Ours-A}                                                                      & \textbf{252.44}    & \textbf{63.96}     & \textbf{350.77}  & \textbf{1.631}      & \textbf{67.04}                           \\
                                                \rowcolor{dark-green!15}      Reduced                    &  \textbf{Ours-B}                               & \textbf{216.02}    & \textbf{54.49}     & \textbf{298.75}  & \textbf{1.395}      & \textbf{66.74}                            \\
                                                 \rowcolor{dark-green!15}                          &  \textbf{Ours-C}                               & \textbf{169.21}    & \textbf{41.13}     & \textbf{238.76}  & \textbf{1.095}      & \textbf{66.45}              \\ \Xhline{3\arrayrulewidth}             
\end{tabular}} 
\label{tab:alg-exp-tiny}
\end{table}
\paragraph{\textbf{{Total Time Costs Comparisons}}}
{We summarized our time costs regarding both supernet-training and search in Table \ref{table: time and efficiency comparison} and compared them with other NAS baselines, including multiplication-based AlphaNet \cite{AlphaNet} and PreNAS \cite{wang2023prenas}, as well as multiplication-reduced NASA-F \cite{nasa-f}}. 

{(i) For \textit{\underline{\textbf{supernet training}}}, compared to \textbf{\textit{multiplication-based baselines}}, such as AlphaNet \cite{AlphaNet} and PreNAS \cite{wang2023prenas}, NASH takes a longer training time.
This is mainly attributed to the slower training and inference speed of customized CUDA kernels for shift and adder layers compared to the PyTorch implementation for multiplications on GPUs \cite{nasa-f, you2023shiftaddvit}. It is worth noting that this time cost overhead is essential for obtaining more hardware-friendly multiplication-reduced hybrid models. Particularly, despite integrating multiplication-free layers, we can achieve comparable accuracy against our most competitive baseline PreNAS \cite{wang2023prenas}. Compared to the \textbf{\textit{multiplication-reduced baseline}} NASA-F \cite{nasa-f}, we have similar training costs due to our similar training recipe. However, thanks to our proposed zero-shot metric utilized to pre-identify promising sub-networks, thus enabling a more advanced concentrated optimization during the subsequent supernet training, we can achieve higher accuracy, i.e., an average of $\uparrow$$0.44\%$ accuracy on CIFAR100, further demonstrating our effectiveness.}

{(ii) For \textit{\underline{\textbf{{architecture search}}}}, due to the utilization of zero-shot metrics, both our NASH and PreNAS \cite{wang2023prenas} can significantly reduce the search costs, when compared with methods that require real accuracy evaluation during network search (i.e., AlphaNet \cite{AlphaNet} and NASA-F \cite{nasa-f}), verifying our search efficiency. Furthermore, to boost hardware efficiency, we integrate \textit{\underline{\textbf{{hardware search}}}} with network search, thus obtaining higher throughput. For example, we can achieve an average $\uparrow$$45\%$ throughput against PreNAS on CIFAR100. }

\subsection{Evaluation of Our Zero-Shot Architecture Search}
\label{sec:exp_alg}

\paragraph{\textbf{Results and Analysis}}
% The models searched by NASS-H are better at the tradeoffs between accuracy and computation cost, as illustrated in Table \ref{}. Specifically, from this table, we conclude that (i) when compared to multiplication-based models, the models searched by NAAS-H incorporate more hardware-friendly multiplication-free operations, thus the computation energy is significantly saved. Meanwhile, the accuracy of ours is still comparable and even better, with $?$, $?$, and $?$ accuracy improvement over multiplication-based PreNAS on CIFAR10, CIFAR100, and Tiny-ImageNet respectively. These advantages over multiplication-based models validate the superiority of the multiplication-reduced method which marries the benefits
% of both multiplication-based convolutions and multiplication-free operations. 
\shh{As presented in Tables \ref{tab:alg-exp-cifar} and \ref{tab:alg-exp-tiny}, the models identified by our NASH framework demonstrate superior trade-offs between accuracy and computational cost. This substantiates our hypothesis that multiplication-reduced models are highly desired to marry the advantages of both convolutions and multiplication-free operators. Furthermore, this also underscores the effectiveness of our proposed zero-shot search in predicting promising multiplication-reduced models to boost accuracy.
Specifically, (i) when compared with \textbf{\textit{multiplication-based models}}, due to the incorporation of hardware-friendly multiplication-free operations, we can significantly save computational energy. Specifically, our energy reductions are $15.5\%$$\sim$$24.7\%$, $3.8\%$$\sim$$17.5\%$, and $17.3\%$$\sim$$23.5\%$ compared to multiplication-based models searched by PreNAS \cite{wang2023prenas} on CIFAR10, CIFAR100, and Tiny-ImageNet, respectively. 
Meanwhile, owing to the effectiveness of our zero-shot search strategy and preference-biased one-shot supernet-based training, we can maintain comparable and even better accuracy.
Particularly, we can offer $-0.07\%$$\sim$$0.25\%$ accuracy improvements over PreNAS \cite{wang2023prenas}.}

\begin{table}[t]
\centering
\caption{The resource consumption of different computation types when implemented on the Kria KV260 FPGA and synthesized under the frequency of 200MHz}
\setlength{\tabcolsep}{8pt}
\resizebox{0.7\linewidth}{!}{
\begin{tabular}{c|cc}
\Xhline{3\arrayrulewidth}
\multirow{2}{*}{\textbf{Computation}} & \multicolumn{2}{c}{\textbf{Resource Consumption}} \\ \cline{2-3} 
                             & \multicolumn{1}{c}{\textbf{LUT}}      & \textbf{DSP}     \\ \hline \hline
\textbf{Conv (Chunk-C)}                        & \multicolumn{1}{c}{37}       & 0.5     \\ \hline
\textbf{Adder (Chunk-A)}                      & \multicolumn{1}{c}{29}       & 0       \\ \hline
\textbf{Shift (Chunk-S)}                    & \multicolumn{1}{c}{34}       & 0       \\ \Xhline{3\arrayrulewidth}
\end{tabular}} \label{tab:resource_comp}
\end{table}

(ii) \shh{As for comparisons with \textbf{\textit{multiplication-free models}}, convolutions within our hybrid models enable us to largely outperform them in terms of accuracy. 
Particularly, we can offer up to $\mathbf{4.2}\%$$\sim$$\mathbf{23.3\%}$ and $\mathbf{6.7}\%$$\sim$$\mathbf{16.6\%}$ higher accuracy over the handcrafted DeepShift-MobileNet-V2 \cite{DeepShift} and AdderNet-MobileNet-V2 \cite{AdderNet}, respectively. 
% Besides, we also achieve $\mathbf{6.7}\%$ and $\mathbf{16.6}\%$ accuracy over AdderNet-MobileNet-V2 \cite{AdderNet} on CIFAR10 and CIFAR100, respectively. 
It is worth noting that although our hybrid models consume more computational energy resulting from the incorporated convolutions, our dedicated chunk-based accelerator that supports the simultaneous processing of heterogeneous layers allows us to achieve comparable and even higher throughput and FPS (as demonstrated in Fig. \ref{fig:overall_results}), when compared with multiplication-free baselines.
}

% Compared to the hand-crafted multiplication-reduced models, our models show significantly enhanced accuracy due to the existence of multiplication-based convolution. To be more precise, we can achieve $?$, $?$ and $?$ accuracy improvement over DeepShift-MobileNet-V2 \cite{DeepShift} on CIFAR10, CIFAR100, and Tiny-ImageNet, while the improvement over AdderNet-MobileNet-V2 \cite{AdderNet} is $?$, $?$ and $?$ on CIFAR10, CIFAR100, and Tiny-ImageNet respectively. It's important to highlight that, despite the higher energy cost because of the increased number of multiplications in our hybrid models compared to the multiplication-free baselines, our multi-chunk accelerator allows us to achieve even better throughput and FPS (as shown in {} and explained in {})

(iii) \shh{Regarding comparisons with hybrid models searched by the SOTA \textbf{\textit{multiplication-reduced system}} NASA-F \cite{nasa-f}, we demonstrate higher accuracy owing to the integration of proposed tailored zero-shot metric.
%advantages of zero-shot NAS over one-shot NAS, as explained in Sec. \ref{sec:intro}. Importantly, zero-shot NAS is applied to multiplication-reduced hybrid models for the first time, thanks to our tailored zero-shot metric proposed in Section \ref{sec:nas}. 
Specifically, we can offer $0.19\%$$\sim$$0.56\%$ higher accuracy. It is noteworthy that despite we gain comparable computational energy costs over NASA-F \cite{nasa-f}, we achieve higher
throughput and FPS (as verified in Fig. \ref{fig:overall_results}). This is attributed to the effectiveness of our coarse-to-fine accelerator search strategy in enhancing hardware efficiency.}

%As for the comparison with the SOTA multiplication-reduced NASA-F system \cite{nasa-f}, while the computation cost is similar, our models can achieve $?$, $?$, and $?$ accuracy improvement on CIFAR10, CIFAR100, and Tiny-ImageNet respectively. This verifies the superiority of zero-shot NAS over one-shot NAS, i.e., we can identify high-quality sub-networks before training and thus training resources can be concentrated to obtain high accuracy. Moreover, despite the similar computation energy cost, NAAS-H can achieve higher throughput and FPS thanks to the more thoroughly designed accelerator (as shown in {}).

\begin{table}[]
\centering
\setlength{\tabcolsep}{3pt}
\caption{Comparisons among models identified by different zero-shot metrics in terms of operation numbers, {computational energy costs (\texttt{Comp.})}, and accuracy (\texttt{Acc.}) {on CIFAR100}} 
\renewcommand{\arraystretch}{1.1}
\resizebox{\linewidth}{!}{
\begin{tabular}{c|ccccc}
\Xhline{3\arrayrulewidth}
\multicolumn{1}{c|}{\multirow{2}{*}{\textbf{Methods}}} & \multicolumn{5}{c}{\textbf{CIFAR100}}                                                                                                                           \\ \cline{2-6} 
\multicolumn{1}{c|}{}                         & \multicolumn{1}{c}{\textbf{Multi. (M)}} & \multicolumn{1}{c}{\textbf{Shift (M)}} & \multicolumn{1}{c}{\textbf{Add (M)}} & \multicolumn{1}{c}{\textbf{Comp. (mJ)}} & \multicolumn{1}{c}{\textbf{Acc. (\%)}} \\ \hline \hline
SNIP \cite{lee2018snip}                                                   &69.14              & 16.10               &106.01            &0.453                &79.02              \\
NN-Degree \cite{nn_degree}                                                   &65.09              & 18.86               &94.03            &0.425                &79.39              \\
Zen-Score \cite{zen-nas}                                                   &66.39              & 16.34               &99.29            &0.434                &79.21              \\
\rowcolor{dark-green!15} \textbf{Ours-C}                                                                      & \textbf{56.61}    & \textbf{14.14}     & \textbf{88.52}  & \textbf{0.373}      & \textbf{79.63}                           \\ \Xhline{3\arrayrulewidth}             
\end{tabular}} \label{tab:alg-exp-abla}
\end{table}
\paragraph{\textbf{Ablation Studies of Our Tailored Zero-Shot Metric}}
% \textcolor{brown}{To validate our tailored zero-shot metric proposed in Sec. \ref{sec:nas}, we replace it with SNIP \cite{lee2018snip}, NN-Degree \cite{nn_degree} and Zen-Score \cite{zen-nas} respectively, and repeat our search-training process on CIFAR100.} These metrics demonstrate weaker correlations with real model accuracy on our multiplication-reduced hybrid models, making it more challenging to identify promising hybrid models. The most competitive model found by each method is presented in Table \ref{}. We observe that models searched by these metrics generally exhibit reduced accuracy after supernet training. It is worth noting that although Zen-Score and NN-Degree are relatively effective compared to SNIP, our tailored zero-shot metric outperforms them by simultaneously taking the trainability and expressivity into consideration.
{To verify our tailored zero-shot metric, we compare it with three SOTA baselines, including SNIP \cite{lee2018snip}, NN-Degree \cite{nn_degree} and Zen-Score \cite{zen-nas}.
As shown in Table \ref{tab:alg-exp-abla}, (i) sub-network identified by our proposed metric achieves the highest accuracy {with much fewer computational energy costs} after the supernet training on {CIFAR100}, demonstrating our effectiveness in predicting promising multiplication-reduced hybrid models to alleviate gradient conflicts and boost accuracy.
{Specifically, we gain $\uparrow$$0.61\%$, $\uparrow$$0.24\%$, and $\uparrow$$0.42\%$ accuracy with $\downarrow$$0.82\times$, 
$\downarrow$$0.88\times$, and $\downarrow$$0.86\times$ computational energy costs compared to SNIP, NN-Degree, and Zen-Score, respectively.}
Besides, (ii) while Zen-Score and NN-Degree individually surpass the gradient-based SNIP by {$\uparrow$$0.37\%$ and $\uparrow$$0.19\%$ accuracy} with fewer computational costs, our tailored metric further enhances accuracy by combining both, which allows us to access both expressivity and trainability of models. 
}

% \usepackage{multirow}
\begin{table*}[t]
\centering
\setlength{\tabcolsep}{1.5pt}
\caption{Comparisons regarding hardware resource consumption, throughput (\texttt{Thrpt.}), FPS, and accuracy, when evaluated on CIFAR10} 
\renewcommand{\arraystretch}{1.2}
\resizebox{\linewidth}{!}{
\begin{threeparttable}{
\begin{tabular}{c||cccc|cc|ccccc}
\Xhline{3\arrayrulewidth}
\multirow{2}{*}{\textbf{Method}} & \multicolumn{4}{c|}{\textbf{Multiplication-Based}}                                      & \multicolumn{2}{c|}{\textbf{Multiplication-Free}}                & \multicolumn{5}{c}{\textbf{Multiplication-Reduced}}                 \\ \cline{2-12} 
                                 & \textbf{AlphaNet} \cite{AlphaNet}      & \textbf{PreNAS-A} \cite{wang2023prenas}         & \textbf{PreNAS-B} \cite{wang2023prenas}    & \textbf{PreNAS-A-C} \cite{wang2023prenas}        & \textbf{DeepShift-MV2} \cite{DeepShift}              & \textbf{AdderNet-MV2} \cite{AdderNet}              & \textbf{NASA-F-A} \cite{nasa-f} & \textbf{NASA-F-B} \cite{nasa-f}           & \color{dark-green}\textbf{Ours-A} & \color{dark-green}\textbf{Ours-B} & \color{dark-green}\textbf{Ours-C} \\ \hline \hline
\textbf{Bitwidth}                & \textbf{INT8}          & \textbf{INT8}          & \textbf{INT8}          & \textbf{INT8}          & \textbf{INT8}                   & \textbf{INT8}               & \textbf{INT8}              & \textbf{INT8}          & \textbf{INT8}            & \textbf{INT8}            & \textbf{INT8}            \\ \hline 
\textbf{\textbf{Strategy}}                & \multicolumn{4}{c|}{\textbf{DSP Implementations}}                                         & \multicolumn{2}{c|}{\textbf{LUT Implementations}} & \multicolumn{5}{c}{\textbf{DSP + LUT Implementations}}               \\ \hline
\textbf{kLUT}                   & 41.3 (35.32\%)          & 41.6 (35.54\%)            & 41.3 (35.32\%)              & 41.1 (35.11\%)              & 46.2 (39.45\%)                                & 46.7 (39.88\%)                               & 63.9 (54.60\%)                            & 80.1 (68.48\%)             & 61.7 (52.75\%)            & 65.4 (55.94\%)            & 66.4 (56.73\%)           \\
\textbf{DSP}                     & 545 (43.7\%) & 545 (43.7\%) & 545 (43.7\%) & 545 (43.7\%)  & --                  & --                & 545 (43.7\%) & 545 (43.7\%)      & 545 (43.7\%) & 545 (43.7\%)   & 545 (43.7\%)   \\
\textbf{BRAM}                    & 80 (27.7\%)   & 80 (27.7\%)   & 80 (27.7\%)   & 80 (27.7\%)   & 80 (27.7\%)   & 80 (27.7\%)         & 80 (27.7\%)       & 80 (27.7\%)        & 25.8 (8.9\%)   & 19.4 (6.7\%)   & 19.2 (6.7\%)     \\
\textbf{Freq. (MHz)}              & 200           & 200           & 200           & 200           & 200           & 200          & 200                 & 200               & 200           & 200                          & 200             \\ \hline
\textbf{Latency (ms)}            & 1.08          & 1.28          & 1.04           & 0.89            & 0.66                                & 0.66                               & 0.72                            & 0.63              & \cellcolor{n3!50}{0.59}            & \cellcolor{n2!50}{0.50}            & \cellcolor{n1!50}{0.44}   \\
\textbf{Thrpt. (GOPS)}           & 134.9         & 133.9             & 133.8             & 136.1             & 258.4                               & 258.4                              & 228.9                          & 308.3             & \cellcolor{n3!50}{315.7}           & \cellcolor{n2!50}{336.7}           & \cellcolor{n1!50}{357.5}  \\
\textbf{GOPS/kLUT}               & 3.265         & 3.219             & 3.237             & 3.313             & \cellcolor{n1!50}{5.598}                               & \cellcolor{n2!50}{5.538}                              & 3.583                         & 3.848             & 5.115           & 5.145           & \cellcolor{n3!50}{5.386}  \\
\textbf{GOPS/DSP}                 & 0.248         & 0.246             & 0.245             & 0.250             & --                                  & --                                 & 0.420                          & 0.566             & \cellcolor{n3!50}{0.579}           & \cellcolor{n2!50}{0.618}           & \cellcolor{n1!50}{0.656}  \\
\textbf{FPS}                     & 925.9         & 783.2             & 957.6             & 1127.1            & 1515.2                              & 1515.2                             & 1388.9                       & 1587.3            & \cellcolor{n3!50}{1706.2}          & \cellcolor{n2!50}{1980.5}          & \cellcolor{n1!50}{2273.6} \\  \hline
\textbf{HW Search}           & {\color{purple}{\xmark}}     & {\color{purple}{\xmark}}              & {\color{purple}{\xmark}}                   & {\color{purple}{\xmark}}                    & {\color{purple}{\xmark}}                    & {\color{purple}{\xmark}}                      & {\color{purple}{\xmark}}                      &  {\color{purple}{\xmark}}            & {\color{dark-green}{\cmark}}                &{\color{dark-green}{\cmark}}               & {\color{dark-green}{\cmark}}                \\ \hline
\textbf{Top-1 Acc. (\%)}                 & 95.86         & \cellcolor{n1!50}{96.19}             & 96.00              & 95.82             & 91.90                               & 89.50                              & 95.96                         & 95.87             & \cellcolor{n2!50}{96.16}           & \cellcolor{n3!50}{96.06}           & 95.87  \\ \Xhline{3\arrayrulewidth}
\end{tabular}}
\end{threeparttable}} \label{tab:hw-exp-cifar10}
\end{table*}

% \usepackage{multirow}
\begin{table*}[t]
\centering
\setlength{\tabcolsep}{1.5pt}
\caption{Comparisons regarding hardware resource consumption, throughput (\texttt{Thrpt.}), FPS, and accuracy, when evaluated on CIFAR100} 
\renewcommand{\arraystretch}{1.2}
\resizebox{\linewidth}{!}{
\begin{threeparttable}{
\begin{tabular}{c||cccc|cc|ccccc}
\Xhline{3\arrayrulewidth}
\multirow{2}{*}{\textbf{Method}} & \multicolumn{4}{c|}{\textbf{Multiplication-Based}}                                      & \multicolumn{2}{c|}{\textbf{Multiplication-Free}}                & \multicolumn{5}{c}{\textbf{Multiplication-Reduced}}                 \\ \cline{2-12} 
                                 & \textbf{AlphaNet} \cite{AlphaNet}      & \textbf{PreNAS-A} \cite{wang2023prenas}         & \textbf{PreNAS-B} \cite{wang2023prenas}    & \textbf{PreNAS-A-C} \cite{wang2023prenas}        & \textbf{DeepShift-MV2} \cite{DeepShift}              & \textbf{AdderNet-MV2} \cite{AdderNet}              & \textbf{NASA-F-A} \cite{nasa-f} & \textbf{NASA-F-B} \cite{nasa-f}           & \color{dark-green}\textbf{Ours-A} & \color{dark-green}\textbf{Ours-B} & \color{dark-green}\textbf{Ours-C} \\ \hline \hline
\textbf{Bitwidth}                & \textbf{INT8}          & \textbf{INT8}          & \textbf{INT8}                 & \textbf{INT8}          & \textbf{INT8}         & \textbf{INT8}               & \textbf{INT8}              & \textbf{INT8}          & \textbf{INT8}            & \textbf{INT8}            & \textbf{INT8}            \\ \hline 
\textbf{\textbf{Strategy}}                & \multicolumn{4}{c|}{\textbf{DSP Implementations}}                                         & \multicolumn{2}{c|}{\textbf{LUT Implementations}} & \multicolumn{5}{c}{\textbf{DSP + LUT Implementations}}               \\ \hline
\textbf{kLUT}                   & 41.8 (35.71\%)          & 42.4 (36.22\%)              & 41.8 (35.71\%)              & 41.4 (35.37\%)              & 47.2 (40.31\%)                                & 48.2 (41.16\%)                               & 53.4 (45.63\%)                          & 58.8 (50.26\%)              & 62.3 (53.21\%)            & 59.5 (50.82\%)            & 59.5 (50.82\%)           \\
\textbf{DSP}                     & 545 (43.7\%) & 545 (43.7\%) & 545 (43.7\%) & 545 (43.7\%)  & --                  & --                & 545 (43.7\%) & 545 (43.7\%)    & 545 (43.7\%) & 545 (43.7\%)   & 545 (43.7\%)   \\
\textbf{BRAM}                    & 80 (27.7\%)   & 80 (27.7\%)   & 80 (27.7\%)   & 80 (27.7\%)   & 80 (27.7\%)   & 80 (27.7\%)         & 80 (27.7\%)          & 80 (27.7\%)     & 39.0 (13.5\%)   & 26.5 (9.2\%)   & 19.2 (6.7\%)     \\
\textbf{Freq. (MHz)}              & 200           & 200           & 200           & 200           & 200           & 200          & 200                 & 200               & 200                      & 200             & 200             \\ \hline
\textbf{Latency (ms)}            & 1.03          & 1.37           & 1.03           & 0.91            & \cellcolor{n3!50}{0.66}                                & \cellcolor{n3!50}{0.66}                               & 0.99                           & 0.74              & 0.68            & \cellcolor{n2!50}{0.63}            & \cellcolor{n1!50}{0.53}   \\
\textbf{Thrpt. (GOPS)}           & 147.6         & 138.6             & 147.1             & 143.1             & 258.4                               & 258.4                              & 197.5                        & 215.9             & \cellcolor{n3!50}{296.6}           & \cellcolor{n1!50}{302.4}           & \cellcolor{n2!50}{299.6}  \\
\textbf{GOPS/kLUT}               & 3.533         & 3.271             & 3.520             & 3.459             & \cellcolor{n1!50}{5.479}                               & \cellcolor{n2!50}{5.366}                              & 3.699                        & 3.672             & 4.764           & \cellcolor{n3!50}{5.086}           & 5.038  \\
\textbf{GOPS/DSP}                 & 0.271         & 0.254             & 0.270             & 0.263             & ---                                 & --                                 & 0.362                          & 0.396             & \cellcolor{n3!50}{0.544}           & \cellcolor{n1!50}{0.555}           & \cellcolor{n2!50}{0.550}  \\
\textbf{FPS}                     & 970.9         & 731.0             & 968.8             & 1102.4            & \cellcolor{n3!50}{1515.2}                              & \cellcolor{n3!50}{1515.2}                             & 1010.1                        & 1351.4            & 1467.5          & \cellcolor{n2!50}{1588.4}          & \cellcolor{n1!50}{1880.9} \\ \hline
\textbf{HW Search}           & {\color{purple}{\xmark}}     & {\color{purple}{\xmark}}              & {\color{purple}{\xmark}}                   & {\color{purple}{\xmark}}                    & {\color{purple}{\xmark}}                    & {\color{purple}{\xmark}}                      & {\color{purple}{\xmark}}                       &  {\color{purple}{\xmark}}            & {\color{dark-green}{\cmark}}                &{\color{dark-green}{\cmark}}               & {\color{dark-green}{\cmark}}                \\ \hline
\textbf{Top-1 Acc. (\%)}                & 79.1          &\cellcolor{n2!50}{80.01}             & 79.80              & 79.38             & 71.00                               & 63.50                              & {79.68}                         & 79.17             & \cellcolor{n1!50}{80.09}           &\cellcolor{n3!50}{79.85}           & 79.63  \\ \Xhline{3\arrayrulewidth}
\end{tabular}}
\end{threeparttable}} \label{tab:hw-exp-cifar100}
\end{table*}
% \usepackage{multirow}
\begin{table*}[t]
\centering
\setlength{\tabcolsep}{2.3pt}
\caption{{Comparisons regarding hardware resource consumption, throughput (\texttt{Thrpt.}), FPS, and accuracy on Tiny-ImageNet}} 
\renewcommand{\arraystretch}{1.2}
\resizebox{\linewidth}{!}{
\begin{tabular}{c||cccc|cc|ccccc}
\Xhline{3\arrayrulewidth}
\multirow{2}{*}{\textbf{Method}} & \multicolumn{4}{c|}{\textbf{Multiplication-Based}}                         & \multicolumn{2}{c|}{\textbf{Multiplication-Free}}                                & \multicolumn{5}{c}{\textbf{Multiplication-Reduced}}                   \\ \cline{2-12} 
                                 & \textbf{AlphaNet} \cite{AlphaNet} & \textbf{PreNAS-A} \cite{wang2023prenas} & \textbf{PreNAS-B} \cite{wang2023prenas} & \textbf{PreNAS-C} \cite{wang2023prenas}  & \textbf{AdderNet-ResNet-20} \cite{AdderNet}   & \textbf{DeepShift-MV2} \cite{DeepShift}  & \textbf{NASA-F-A} \cite{nasa-f} & \textbf{NASA-F-B} \cite{nasa-f}  & \color{dark-green}{\textbf{Ours-A}} & \color{dark-green}{\textbf{Ours-B}} & \color{dark-green}{\textbf{Ours-C}}  \\ \hline \hline
\textbf{Bitwidth}                    & \textbf{INT8}       & \textbf{INT8}       & \textbf{INT8}     & \textbf{INT8}       & \textbf{INT8}   & \textbf{INT8}       & \textbf{INT8}   & \textbf{INT8}   & \textbf{INT8}   & \textbf{INT8}   & \textbf{INT8}   \\ \hline
\textbf{Strategy}                & \multicolumn{4}{c|}{\textbf{DSP Implementations}}                     & \multicolumn{2}{c|}{\textbf{LUT Implementations}}                               & \multicolumn{5}{c}{\textbf{DSP + LUT Implementations}}        \\ \hline
\textbf{kLUT}                    & 42.9 (36.65\%)          & 43.1 (36.82\%)              & 42.0 (35.88\%)              & 41.7 (35.62\%)              & 48.3 (41.28\%)                                & 60.9 (52.05\%)                               & 52.6 (44.97\%)              & 67.7 (57.89\%)                           & 54.5 (46.57\%)            & 55.2 (47.17\%)            & 56.3 (48.16\%)            \\
\textbf{DSP}                     & 545 (43.67\%) & 545 (43.67\%)     & 545 (43.67\%)     & 545 (43.67\%)     & --                                  & --                                      & 545 (43.67\%)     & 545 (43.67\%)     & 545 (43.67\%)   & 545 (43.67\%)   & 545 (43.67\%)   \\
\textbf{BRAM}                    & 80 (27.7\%)   & 80 (27.7\%)       & 80 (27.7\%)       & 80 (27.7\%)       & 80 (27.7\%)                         & 80 (27.7\%)                               & 80 (27.7\%)       & 80 (27.7\%)       & 52.5 (18.2\%)     & 44.8 (15.5\%)      & 73.0 (25.3\%)     \\
\textbf{Freq.(MHz)}              & 200           & 200               & 200               & 200               & 200                                 & 200                                & 200               & 200                              & 200             & 200             & 200            \\ \hline
\textbf{Latency (ms)}          & 2.98          & 3.34              & 2.77              & 2.41              & \cellcolor{n3!50}{1.4}                                 & \cellcolor{n1!50}{1.04}                               & 2                 & 1.65                            & 1.68            & 1.42            & \cellcolor{n2!50}{1.11}            \\
\textbf{Thrpt. (GOPS)}           & 215.1         & 205.1             & 203.0             & 198.0              & 335.4                               & 368.9                              & 282.4             & 354.7                          & \cellcolor{n3!50}{396.7}           & \cellcolor{n2!50}{400.1}           & \cellcolor{n1!50}{405.6}           \\ 
\textbf{GOPS/kLUT}                & 5.015         & 4.760             & 4.836             & 4.751             & 6.945                               & 6.058                              & 5.368             & 5.237                          & \cellcolor{n1!50}{7.280}           & \cellcolor{n2!50}{7.249}           & \cellcolor{n3!50}{7.197}          \\
\textbf{GOPS/DSP}                & 0.395         & 0.376             & 0.372             & 0.152             & --                                  & --                                 & 0.518             & 0.651                          & \cellcolor{n3!50}{0.728}           & \cellcolor{n2!50}{0.734}           & \cellcolor{n1!50}{0.744}          \\
\textbf{FPS}                     & 335.6         & 299.4             & 361.0             & 414.9             & \cellcolor{n3!50}{714.3}                               & \cellcolor{n1!50}{961.5}                              & 500.0             & 606.1                        & 594.6           & 702.8           & \cellcolor{n2!50}{903.0}           \\ \hline
\textbf{HW Search}           & {\color{purple}{\xmark}}     & {\color{purple}{\xmark}}              & {\color{purple}{\xmark}}                   & {\color{purple}{\xmark}}                    & {\color{purple}{\xmark}}                    & {\color{purple}{\xmark}}                      & {\color{purple}{\xmark}}                       &  {\color{purple}{\xmark}}            & {\color{dark-green}{\cmark}}                &{\color{dark-green}{\cmark}}               & {\color{dark-green}{\cmark}}                \\ \hline
\textbf{Top-1 Acc. (\%)}               & 65.99         & \cellcolor{n2!50}{66.88}             & 66.63             & 66.49             & 55.3                                & 43.7                               & 66.51             & 66.18                          & \cellcolor{n1!50}{67.04}           & \cellcolor{n3!50}{66.74}           & {66.45}           \\ \Xhline{3\arrayrulewidth}
\end{tabular}} \label{table:hw-exp-tiny} 
\end{table*}

\subsection{Evaluation of Coarse-to-Fine Accelerator Search}
\label{sec:exp_hw}

\paragraph{\textbf{Results and Analysis}}
% \textcolor{brown}{As illustrated in Tables \ref{tab:hw-exp-cifar10}, \ref{tab:hw-exp-cifar100} and \ref{table:hw-exp-tiny}, where the \fboxsep1.5pt\colorbox{n1!50}{first}, \fboxsep1.5pt\colorbox{n2!50}{second}, and \fboxsep1.5pt\colorbox{n3!50}{third} ranking performances are noted with corresponding colors, the accelerators searched by our coarse-to-fine accelerator search strategy (as introduced in Sec. \ref{sec:accelerator search}) provide our hybrid models with enhanced hardware efficiency (i.e., the overall throughput, hardware utilization efficiency w.r.t. both DSP slices and LUT resources, and FPS). This effectiveness validates our accelerator design, which strategically utilizes heterogeneous FPGA resources to construct customized chunks, resulting in enhanced throughput and hardware utilization. Besides, this also underscores the efficacy of our coarse-to-fine search which fully explores the hardware design opportunities. To be more specific, (i) when compared with \textbf{\textit{multiplication-based models}} on their dedicated accelerators, the throughput along with FPS is greatly improved (as illustrated in Fig. \ref{fig:overall_results}) since we leverage LUTs to build multiple chunks simultaneously processing heterogeneous computations. Moreover, the efficiency of DSP slices and LUTs is also significantly enhanced because of the improved throughput. Precisely, we can achieve up to $?$, $?$, and $?$ DSP slices efficiency improvement as well as $?$, $?$, and $?$ higher LUTs efficiency on CIFAR10, CIFAR100, and Tiny-ImageNet respectively.}
\shh{As illustrated in Tables \ref{tab:hw-exp-cifar10}, \ref{tab:hw-exp-cifar100} and \ref{table:hw-exp-tiny}, where the \fboxsep1.5pt\colorbox{n1!50}{first}, \fboxsep1.5pt\colorbox{n2!50}{second}, and \fboxsep1.5pt\colorbox{n3!50}{third} ranking performances are noted with corresponding colors, 
% dedicated accelerators searched via our coarse-to-fine accelerator search strategy (as introduced in Sec. \ref{sec:accelerator search}) 
we achieve superior hardware efficiency (i.e., the overall throughput, hardware utilization efficiency w.r.t. both DSPs and LUTs, and FPS). This effectiveness highlights our chunk-based design, which utilizes heterogeneous computing resources available on FPGAs to construct customized chunks with enhanced hardware utilization and improved throughput. Besides, this also underscores the efficacy of our coarse-to-fine accelerator search strategy in fully unleashing hardware acceleration opportunities. Specifically, (i) when compared with \textbf{\textit{multiplication-based baselines}}, due to the utilization of heterogeneous computing resources, which yields enhanced utilization efficiency in DSPs and LUTs, our throughput and FPS are greatly improved. Particularly, we can achieve up to $\uparrow$$\mathbf{1.63\times}$, $\uparrow$$\mathbf{1.46\times}$, and $\uparrow$$\mathbf{1.53\times}$ DSP utilization efficiency alongside $\uparrow$$\mathbf{2.63\times}$, $\uparrow$$\mathbf{2.14\times}$, and $\uparrow$$\mathbf{2.05\times}$ LUT utlization efficiency on CIFAR10, CIFAR100, and Tiny-ImageNet, respectively.}

% (ii) \textcolor{brown}{The dedicated accelerators for \textbf{\textit{multiplication-free models}} rely on the LUTs to process hardware-friendly additions or bitwise shifts, thus obtaining excellent throughput and FPS. However, thanks to our coarse-to-fine accelerator search which thoroughly designs the accelerators for the models from NAAS-H, we can offer comparable and even better throughput and FPS performance when compared to multiplication-free baselines. However, we do not sacrifice models' accuracy to achieve this outstanding throughput and FPS, as shown in Fig. \ref{fig:overall_results}.
% }

\shh{(ii) In comparisons to \textbf{\textit{multiplication-free baselines}}, which solely rely on LUTs to process additions or bitwise shifts, thus obtaining the highest LUT utilization efficiency, we can offer comparable and even better throughput and FPS with much higher accuracy. This is attributed to our chunk-based design that simultaneously processes heterogeneous layers within our hybrid models with distinct customized chunks.}

{More specifically, we have lower \underline{\textit{\textbf{GOPS/kLUT}}} on CIFAR10 and CIFAR100, and higher GOPS/kLUT on Tiny-ImageNet. The lower performance on CIFAR10 and CIFAR100 is mainly due to \textbf{\textit{higher LUT consumption}} by the convolution-dedicated computing engine (Chunk-C) compared to the multiplication-free adder (Chunk-A) and shift layers (Chunk-S), as shown in Table \ref{tab:resource_comp}. This higher LUT consumption in Chunk-C stems from three factors. {{{1)}}} We follow NASA-F \cite{nasa-f} to employ the SOTA DSP-packing strategy \cite{Xilinx-conv} to pack two $8$-bit multiplications within one DSP, which enhances DSP utilization efficiency but yields additional LUT consumption. 
{{{2)}}} Higher bit-width representations in convolutions require larger registers (e.g., $8$-bit weights and $15$-bit outputs in convolutions vs. $4$-bit weights in shift layers and $9$-bit outputs in adder layers), which leads to increased LUT consumption.
{{{(iii)}}} Larger output bit-widths also necessitate bigger accumulators, further increasing LUT usage.
Despite the higher LUT consumption in Chunk-C, our models searched on Tiny-ImageNet are significantly larger compared to handcraft multiplication-based models. The larger model sizes not only improve accuracy but also enhance hardware utilization, leading to improved GOPS/kLUT on this challenging dataset.}

{For \underline{\textit{\textbf{FPS}}}, compared to multiplication-free baselines, we achieve higher FPS on CIFAR10, comparable FPS on CIFAR100, and lower FPS on Tiny-ImageNet. This variation in performance is primarily influenced by \textit{\textbf{operation numbers}}. Contrary to handcrafted multiplication-free models, which use the same architecture across all datasets and thus inevitably yield accuracy degradation on more challenging datasets, our NAS algorithm automatically searches for models of varying sizes tailored to each dataset to preserve accuracy, thereby resulting in different FPS outcomes.}

{Note that in situations where our hardware efficiency is lower compared to multiplication-free baselines, our superiority primarily lies in significantly \textbf{\textit{higher accuracy}}, as outlined in Secs. \ref{sec:oversota} and \ref{sec:exp_alg}.
Despite the promising hardware efficiency of multiplication-free baselines, with the integration of our dedicated accelerator, we can provide comparable or even superior hardware efficiency in certain scenarios, as discussed in the above paragraphs.}
% (iii) \textcolor{brown}{We also outperform the SOTA 
% \textbf{\textit{multiplication-reduced system}} NASA-F \cite{nasa-f} on its dedicated FPGA accelerators through our coarse-to-fine accelerator search. Besides the improved throughput and FPS shown in Fig. \ref{fig:overall_results}, the efficiency of DSP slices and LUTs is also enhanced since we avoid the additional hardware resource consumption in our accelerator search, which is introduced in Sec. \ref{sec:c2f}. More specifically, we achieve up to $?$, $?$, and $?$ higher DSP slices efficiency as well as $?$, $?$, and $?$ LUTs efficiency improvement when tested on CIFAR10, CIFAR100, and Tiny-ImageNet respectively
% }

\shh{(iii) When compared with the SOTA \textbf{\textit{multiplication-reduced baseline}} NASA-F \cite{nasa-f}, owing to our proposed coarse-to-fine accelerator search strategy, we can gain better hardware efficiency (i.e., higher FPS and throughput) with less hardware resource consumption (i.e., higher DSP and LUT utilization efficiency).
For example, we can gain up to $\uparrow$$\mathbf{1.44\times}$, $\uparrow$$\mathbf{1.39\times}$, and $\uparrow$$\mathbf{1.36\times}$ DSP utilization efficiency as well as $\uparrow$$\mathbf{1.47\times}$, $\uparrow$$\mathbf{1.39\times}$, and $\uparrow$$\mathbf{1.42\times}$ LUT utilization efficiency, when tested on CIFAR10, CIFAR100, and Tiny-ImageNet, respectively.}

\paragraph{\textbf{Ablation Studies of Coarse-to-Fine Search Strategy}}
\huihong{To demonstrate the efficiency and accuracy of our coarse-to-fine search strategy, we compare it with three search strategies. As listed in Table \ref{table:hw-abl}: (i) Iteratively exploring the vanilla huge search space; (ii) Solely performing coarse search for Chunk-C and manually designing other components following NASA-F \cite{nasa-f}; (iii) Manually designing Chunk-C following NASA-F \cite{nasa-f} while performing fine search for other components. As verified in Table \ref{table:hw-abl}, when compared with (i), we can achieve comparable throughput with a much faster search speed, demonstrating our effectiveness. As for comparisons with (ii) and (iii), we can gain higher throughput with negligible time overheads, further highlighting our superiority in boosting hardware efficiency while enhancing search efficiency. 
% It is noteworthy that the throughput degradation in (ii) and (iii) indicates the quality of our manual design instead of the importance of the coarse or fine search.
}
\begin{table}[]
\centering
\setlength{\tabcolsep}{6pt}
\caption{Comparisons of search accuracy and search efficiency among various search strategies when tested on CIFAR100. The search time cost and average throughput (avg. thrpt.) of the searched accelerator are provided.} 
\renewcommand{\arraystretch}{1.1}
\resizebox{0.9\linewidth}{!}{
\begin{tabular}{c|c|c|c}
\Xhline{3\arrayrulewidth}
\multicolumn{1}{l|}{\textbf{Coarse Search}} & \multicolumn{1}{l|}{\textbf{Fine Search}} & \multicolumn{1}{l|}{\textbf{Avg. Thrpt. (GOPS)}} & \textbf{Time Cost} \\ \hline \hline
{\color{purple}{\xmark}}                                            &  {\color{purple}{\xmark}}                                          &{299.5}                                               &73 hours                    \\
{\color{dark-green}{\cmark}}                                            & {\color{purple}{\xmark}}                                         &254.0                                            &2 hours                    \\ 
{\color{purple}{\xmark}}                                            & {\color{dark-green}{\cmark}}                                         &292.4                                            &4 hours                    \\ 
\rowcolor{dark-green!15}  {\color{dark-green}{\cmark}}                                            & {\color{dark-green}{\cmark}}                                         &\textbf{299.5}                                           &\textbf{5.5 hours}                    \\ \Xhline{3\arrayrulewidth}
\end{tabular}} \label{table:hw-abl} 
\end{table} 
%\vspace{-0.3em}
\section{Conclusion}
% %\vspace{-0.8em}
This paper introduces, formulates, and validates NASH, a \textbf{N}eural architecture and \textbf{A}ccelerator co-\textbf{S}earch framework dedicated to multiplication-reduced \textbf{H}ybrid models. 
Specifically, regarding \underline{neural architecture search}, we introduce a \textbf{\textit{tailored zero-shot metric}} to pre-identify promising multiplication-reduced sub-architectures before supernet training, thus enhancing search efficiency while boosting accuracy.
In terms of \underline{accelerator search}, we innovatively propose a \textbf{\textit{coarse-to-fine accelerator search strategy}} to expedite the accelerator search process and improve hardware efficiency.
Furthermore, we integrate the neural architecture search with accelerator search to obtain NASH, aiming to directly obtain the optimal pairing of hybrid models and dedicated accelerators.
Extensive experimental results validate our effectiveness. Particularly, {we offer $\uparrow$$0.56\%$ accuracy on Tiny-ImageNet compared to the prior multiplication-reduced work NASA-F, and up to $\uparrow$$2.14\times$ throughput and $\uparrow$$2.01\times$ FPS with $\uparrow$$0.25\%$ accuracy over the state-of-the-art multiplication-based system on CIFAR100.

We will focus on two aspects in the future: (i) Improving the predictive accuracy of the tailored zero-shot metric to enhance accuracy. 
%mark
(ii) Expanding the dataflow search space beyond the typical four dataflows to enable a more comprehensive accelerator search. 
% Currently, given the extensive search space introduced by multiple chunks in our accelerator, we employ four typical dataflows for each chunk to facilitate efficient exploration. Thus, a forthcoming direction includes further expansion of the dataflow search space.

% \bibliographystyle{unsrt}
\bibliographystyle{IEEEtran}
\bibliography{nas}

\end{document}